\documentclass[journal]{IEEEtran}
\usepackage{amsmath,amssymb,amsfonts}
\usepackage{graphicx}
\usepackage{booktabs}
\usepackage{multirow}
\usepackage{cite}
\usepackage{url}

\usepackage[dvipsnames]{xcolor}
\usepackage{tikz}
\usetikzlibrary{arrows.meta,positioning,calc,fit,backgrounds}
\definecolor{pblue}{HTML}{2A78D6}
\definecolor{porange}{HTML}{EB6834}
\definecolor{pviolet}{HTML}{4A3AA7}
\definecolor{pgrey}{HTML}{8A8A86}
\usepackage{array}
\usepackage{orcidlink}
\hypersetup{hidelinks,
  pdftitle={Dynamic State Estimation from Sparse PMUs via Physics-Informed Graph Neural Operators},
  pdfauthor={Huy Hoang Le, Haoguang Wang, Christian Moya, Marcos Netto, and Guang Lin},
  pdfkeywords={Dynamic state estimation, graph neural operator, physics-informed learning, phasor measurement units, model error},
  pdfsubject={IEEE Transactions on Power Systems}}

\newcommand{\xhat}{\hat{x}}
\newcommand{\R}{\mathbb{R}}
\newcommand{\Gcal}{\mathcal{G}}
\newcommand{\Loss}{\mathcal{L}}
\newcommand{\cmk}{\checkmark}
\newcommand{\sigmark}{\rlap{$^{\ast}$}}

\begin{document}

\title{Dynamic State Estimation from Sparse PMUs via\\ Physics-Informed Graph Neural Operators}

\author{Huy~Hoang~Le\,\orcidlink{0009-0009-2138-5468},
        Haoguang~Wang\,\orcidlink{0009-0005-4101-864X},
        Christian~Moya\,\orcidlink{0000-0003-0180-9285},
        Marcos~Netto\,\orcidlink{0000-0001-7002-3345},~\IEEEmembership{Senior Member,~IEEE,}
        and~Guang~Lin\,\orcidlink{0000-0002-0976-1987}%
\thanks{We acknowledge support from the National Science Foundation
(DMS2533878, DMS-2053746, DMS-2134209, ECCS-2328241, CBET-2347401,
OAC-2311848); the Department of Energy Office of Science Advanced
Scientific Computing Research program (DE-SC0023161) and Fusion Energy
Sciences program (DE-SC0024583); and the SciDAC LEADS Institute.}%
\thanks{(Corresponding author: Guang Lin.)}%
\thanks{H.~H.~Le, H.~Wang, and G.~Lin are with the School of Mechanical Engineering,
  Purdue University, West Lafayette, IN~47907~USA
  (e-mail: \mbox{lehuyhoang02012002@gmail.com}; \mbox{wang6411@purdue.edu}; \hspace{0pt plus 1fil}\penalty0\hspace{0pt plus -1fil}\mbox{guanglin@purdue.edu}).}%
\thanks{C.~Moya and G.~Lin are with the Department of Mathematics,
  Purdue University, West Lafayette, IN~47907~USA
  (e-mail: \mbox{cmoyacal@purdue.edu}).}%
\thanks{M.~Netto is with the Department of Electrical and Computer Engineering,
  New Jersey Institute of Technology, Newark, NJ~07102~USA
  (e-mail: \mbox{marcos.netto@njit.edu}).}%
}

\markboth{}{}

\maketitle

\begin{abstract}
This paper proposes a physics-informed graph neural operator for power system dynamic state estimation. From a sparse, intermittent measurement window at a few generator buses, the operator estimates the rotor-angle and speed trajectories of all machines, unmeasured ones included, in one forward pass. A graph-attention encoder processes the measurements, and the swing equations are imposed as a hard constraint or as a soft one with a learned correction that absorbs model errors. It is trained on simulated measurements, without state labels. With several phasor measurement units, masked training hides one from the input but keeps it in the loss. On four transmission systems, simulations show that the operator is the most accurate estimator that neither optimizes per scenario nor uses state labels, except on the smallest system at large model errors, where tuned Kalman smoothers are better. Networks trained on simulated states are more accurate on the two smallest systems but not on the 118-bus system, where the operator is faster and more accurate than hard-constrained moving-horizon estimation. Under model errors, it matches or outperforms an optimization-based benchmark at a fraction of its cost. The results hold for a detailed machine model with exact parameters seen only through voltage phasors.
\end{abstract}

\begin{IEEEkeywords}
Dynamic state estimation, graph neural networks, moving-horizon estimation, phasor measurement units, physics-informed machine learning.
\end{IEEEkeywords}

\section{Introduction}\label{sec:intro}
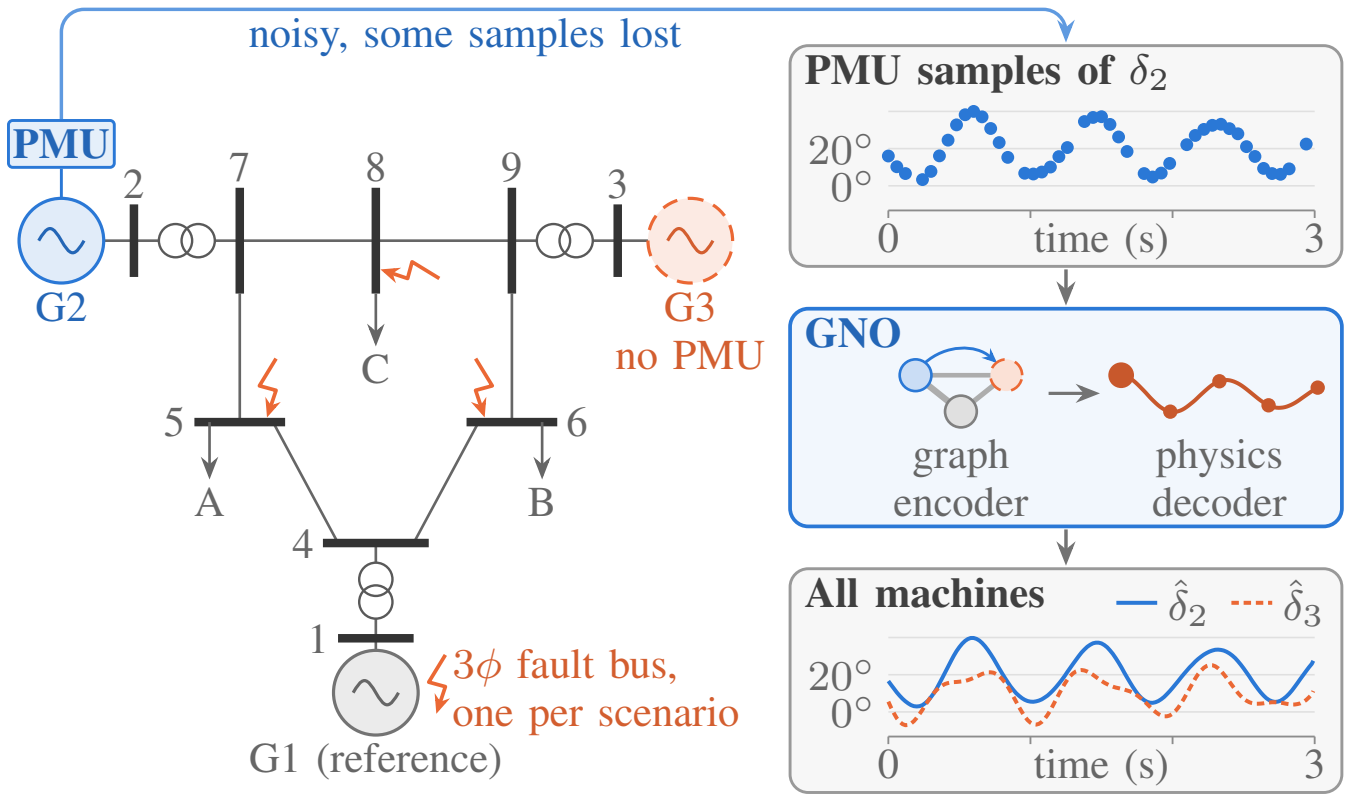
\begin{figure}[t]
\centering
\begin{tikzpicture}[
  x=1mm,y=1mm,
  >={Stealth[length=1.5mm,width=1.15mm]},
  font=\footnotesize,
  bar/.style  ={line width=1.6pt,draw=black!80},
  ln/.style   ={line width=0.5pt,draw=black!60},
  xf/.style   ={circle,draw=black!65,line width=0.45pt,minimum size=2.2mm,inner sep=0pt},
  gen/.style  ={circle,line width=0.6pt,minimum size=5.6mm,inner sep=0pt},
  bn/.style   ={font=\footnotesize,text=black!60,inner sep=0.4pt},
  tag/.style  ={font=\footnotesize,inner sep=0.4pt},
  ttl/.style  ={font=\footnotesize\bfseries,text=black!75,inner sep=0pt,anchor=north west},
  pmu/.style  ={draw=pblue,fill=pblue!12,rounded corners=0.8pt,line width=0.55pt,font=\footnotesize\bfseries,
                text=pblue!85!black,inner sep=0.8pt,minimum width=7mm,minimum height=3.1mm},
  gcard/.style={rounded corners=1.2mm,line width=0.55pt,draw=black!40,fill=black!3},
  ccard/.style={rounded corners=1.2mm,line width=0.55pt,draw=#1,fill=#1!6},
  nd/.style   ={circle,inner sep=0pt,minimum size=2.1mm,line width=0.5pt},
  dot/.style  ={circle,inner sep=0pt,minimum size=0.85mm,fill=pblue},
  flow/.style ={->,>={Stealth[length=1.6mm,width=1.25mm]},line width=0.65pt,draw=black!55},
  fault/.style={draw=porange,line width=0.7pt,line join=round,-{Stealth[length=1.25mm,width=1.05mm]}}]
\draw[ln] (6.10,11.50) -- (8.10,11.50);
\draw[ln] (40.10,11.50) -- (42.10,11.50);
\draw[ln] (15.10,11.50) -- (24.10,11.50);
\draw[ln] (24.10,11.50) -- (33.10,11.50);
\draw[ln] (15.10,8.00) -- (15.10,-0.50);
\draw[ln] (33.10,8.00) -- (33.10,-0.50);
\draw[ln] (17.30,-0.50) -- (21.60,-8.50);
\draw[ln] (30.90,-0.50) -- (26.60,-8.50);
\draw[ln] (24.10,-14.80) -- (24.10,-15.60);
\draw[ln] (8.10,11.50) -- (9.75,11.50); \draw[ln] (13.45,11.50) -- (15.10,11.50);
\node[xf] at (10.85,11.50) {}; \node[xf] at (12.35,11.50) {};
\draw[ln] (33.10,11.50) -- (34.75,11.50); \draw[ln] (38.45,11.50) -- (40.10,11.50);
\node[xf] at (35.85,11.50) {}; \node[xf] at (37.35,11.50) {};
\draw[ln] (24.10,-8.50) -- (24.10,-9.80); \draw[ln] (24.10,-13.50) -- (24.10,-14.80);
\node[xf] at (24.10,-10.90) {}; \node[xf] at (24.10,-12.40) {};
\draw[bar] (8.10,9.10) -- (8.10,13.90);
\draw[bar] (15.10,8.00) -- (15.10,15.00);
\draw[bar] (24.10,8.00) -- (24.10,15.00);
\draw[bar] (33.10,8.00) -- (33.10,15.00);
\draw[bar] (40.10,9.10) -- (40.10,13.90);
\draw[bar] (12.10,-0.50) -- (18.10,-0.50);
\draw[bar] (30.10,-0.50) -- (36.10,-0.50);
\draw[bar] (20.60,-8.50) -- (27.60,-8.50);
\draw[bar] (21.60,-14.80) -- (26.60,-14.80);
\node[bn,anchor=south] at (8.10,14.10) {2};
\node[bn,anchor=south] at (15.10,15.20) {7};
\node[bn,anchor=south] at (24.10,15.20) {8};
\node[bn,anchor=south] at (33.10,15.20) {9};
\node[bn,anchor=south] at (40.10,14.10) {3};
\node[bn,anchor=east] at (11.70,-0.50) {5};
\node[bn,anchor=west] at (36.50,-0.50) {6};
\node[bn,anchor=east] at (20.20,-8.50) {4};
\node[bn,anchor=east] at (21.20,-14.80) {1};
\draw[ln,->] (13.10,-0.50) -- (13.10,-4.30); \node[bn,anchor=north] at (13.10,-4.60) {A};
\draw[ln,->] (35.10,-0.50) -- (35.10,-4.30); \node[bn,anchor=north] at (35.10,-4.60) {B};
\draw[ln,->] (24.10,8.00) -- (24.10,4.50); \node[bn,anchor=north] at (24.10,4.20) {C};
\draw[fault] (17.50,3.70) -- (16.40,1.90) -- (17.50,1.60) -- (16.90,-0.20);
\draw[fault] (30.70,3.70) -- (31.80,1.90) -- (30.70,1.60) -- (31.30,-0.20);
\draw[fault] (28.30,9.00) -- (26.50,10.10) -- (26.20,9.00) -- (24.40,9.60);
\draw[fault] (28.70,-15.90) -- (27.60,-17.70) -- (28.70,-18.00) -- (28.10,-19.80);
\node[bn,text=porange!90!black,anchor=north west,align=left,inner sep=0pt] at (29.10,-15.40) {$3\phi$ fault bus,\\one per scenario};
\node[gen,draw=pblue,fill=pblue!14] at (3.30,11.50) {};
\draw[pblue!85!black,line width=0.6pt] (1.80,11.50) sin (2.55,12.30) cos (3.30,11.50) sin (4.05,10.70) cos (4.80,11.50);
\node[gen,draw=porange,densely dashed,fill=porange!14] at (44.90,11.50) {};
\draw[porange!90!black,line width=0.6pt] (43.40,11.50) sin (44.15,12.30) cos (44.90,11.50) sin (45.65,10.70) cos (46.40,11.50);
\node[gen,draw=black!55,fill=black!8] at (24.10,-18.40) {};
\draw[black!60,line width=0.6pt] (22.60,-18.40) sin (23.35,-17.60) cos (24.10,-18.40) sin (24.85,-19.20) cos (25.60,-18.40);
\node[tag,text=pblue!85!black,anchor=north] at (3.30,8.30) {G2};
\node[tag,text=porange!90!black,anchor=north,align=center] at (44.90,8.30) {G3\\no PMU};
\node[tag,text=black!60,anchor=north] at (24.10,-21.50) {G1 (reference)};
\node[pmu] (pmu) at (3.30,17.90) {PMU};
\draw[pblue,line width=0.55pt] (3.30,14.30) -- (pmu.south);
\draw[pblue!75,line width=0.65pt,rounded corners=1.5mm,->] (pmu.north) |- (69.75,26.80) -- (69.75,24.60);
\node[bn,text=pblue!85!black,anchor=north] at (30.0,26.45) {noisy, some samples lost};
\draw[gcard] (51.50,9.80) rectangle (88.00,24.40);
\node[ttl] at (52.40,23.90) {PMU samples of $\delta_2$};
\draw[black!12,line width=0.3pt] (58.00,15.13) -- (86.20,15.13);
\node[bn,anchor=east] at (57.40,15.13) {$0^\circ$};
\draw[black!12,line width=0.3pt] (58.00,17.59) -- (86.20,17.59);
\node[bn,anchor=east] at (57.40,17.59) {$20^\circ$};
\draw[black!12,line width=0.3pt] (58.00,20.05) -- (86.20,20.05);
\draw[black!45,line width=0.35pt] (58.00,13.50) -- (86.20,13.50);
\draw[black!45,line width=0.35pt] (58.00,13.50) -- ++(0,-0.5);
\draw[black!45,line width=0.35pt] (67.40,13.50) -- ++(0,-0.5);
\draw[black!45,line width=0.35pt] (76.80,13.50) -- ++(0,-0.5);
\draw[black!45,line width=0.35pt] (86.20,13.50) -- ++(0,-0.5);
\node[bn,anchor=north] at (58.00,12.90) {0};
\node[bn,anchor=north] at (86.20,12.90) {3};
\node[bn,anchor=north] at (72.10,12.90) {time (s)};
\foreach \p in {(58.00,17.10),(58.56,16.39),(59.13,15.94),(60.26,15.54),(60.82,16.08),(61.38,17.11),(61.95,18.15),(62.51,19.16),(63.08,19.83),(63.64,20.05),(64.20,19.69),(64.77,18.92),(65.33,17.99),(65.90,17.01),(67.02,15.96),(67.59,15.91),(68.15,16.04),(68.72,16.37),(69.28,17.06),(69.84,17.66),(70.97,19.38),(71.54,19.65),(72.10,19.70),(72.66,19.19),(73.23,18.36),(73.79,17.40),(74.92,15.94),(75.48,15.71),(76.05,15.97),(76.61,16.61),(77.74,17.85),(78.30,18.46),(78.87,18.86),(79.43,19.13),(80.00,19.20),(80.56,18.91),(81.12,18.57),(81.69,17.72),(82.25,17.07),(82.82,16.28),(83.38,15.94),(83.94,15.89),(84.51,16.25),(85.64,17.89)} {\node[dot] at \p {};}
\draw[flow] (69.75,9.60) -- (69.75,7.20);
\draw[ccard=pblue,line width=0.7pt] (51.50,-7.40) rectangle (88.00,7.00);
\node[ttl,text=pblue!85!black] at (52.40,6.50) {GNO};
\draw[black!32,line width=1.2pt] (62.80,0.20) -- (59.80,2.60); \draw[black!32,line width=0.97pt] (62.80,0.20) -- (65.80,2.60);
\draw[black!32,line width=0.87pt] (59.80,2.60) -- (65.80,2.60);
\draw[->,pblue,line width=0.55pt,>={Stealth[length=1mm,width=0.85mm]}] (60.40,3.40) to[bend left=42] (65.20,3.40);
\node[nd,draw=pblue,fill=pblue!25] at (59.8,2.6) {};
\node[nd,draw=porange,densely dashed,fill=porange!20] at (65.8,2.6) {};
\node[nd,draw=black!50,fill=black!12] at (62.8,0.2) {};
\node[bn,anchor=north,align=center] at (62.8,-1.4) {graph\\encoder};
\draw[flow,line width=0.6pt] (68.6,1.4) -- (72.0,1.4);
\draw[porange!85!black,line width=0.75pt] plot[smooth] coordinates {(73.40,2.60) (73.67,2.55) (73.94,2.42) (74.21,2.24) (74.48,2.00) (74.75,1.73) (75.03,1.44) (75.30,1.14) (75.57,0.87) (75.84,0.62) (76.11,0.42) (76.38,0.27) (76.65,0.19) (76.92,0.17) (77.19,0.22) (77.46,0.33) (77.73,0.50) (78.00,0.71) (78.28,0.94) (78.55,1.20) (78.82,1.45) (79.09,1.69) (79.36,1.91) (79.63,2.08) (79.90,2.20) (80.17,2.27) (80.44,2.28) (80.71,2.23) (80.98,2.13) (81.25,1.99) (81.53,1.80) (81.80,1.59) (82.07,1.37) (82.34,1.15) (82.61,0.94) (82.88,0.76) (83.15,0.61) (83.42,0.51) (83.69,0.45) (83.96,0.45) (84.23,0.49) (84.50,0.58) (84.78,0.71) (85.05,0.87) (85.32,1.06) (85.59,1.25) (85.86,1.44) (86.13,1.62) (86.40,1.78)};
\node[circle,inner sep=0pt,minimum size=1.7mm,fill=porange!85!black] at (73.40,2.60) {};
\foreach \p in {(76.65,0.19),(79.90,2.20),(83.15,0.61),(86.40,1.78)} {\node[circle,inner sep=0pt,minimum size=0.95mm,fill=porange!85!black] at \p {};}
\node[bn,anchor=north,align=center] at (79.9,-1.4) {physics\\decoder};
\draw[flow] (69.75,-7.60) -- (69.75,-10.00);
\draw[gcard] (51.50,-25.00) rectangle (88.00,-10.20);
\node[ttl] at (52.40,-10.70) {All machines};
\node[bn,anchor=north east] (l3) at (87.10,-10.70) {$\hat\delta_3$};
\draw[porange,line width=0.75pt,dash pattern=on 1.4pt off 0.9pt] ([xshift=-3.2mm]l3.west) -- ([xshift=-0.5mm]l3.west);
\node[bn,anchor=east] (l2) at ([xshift=-4.4mm]l3.west) {$\hat\delta_2$};
\draw[pblue,line width=0.75pt] ([xshift=-3.2mm]l2.west) -- ([xshift=-0.5mm]l2.west);
\draw[black!12,line width=0.3pt] (58.00,-19.67) -- (86.20,-19.67);
\node[bn,anchor=east] at (57.40,-19.67) {$0^\circ$};
\draw[black!12,line width=0.3pt] (58.00,-17.21) -- (86.20,-17.21);
\node[bn,anchor=east] at (57.40,-17.21) {$20^\circ$};
\draw[black!12,line width=0.3pt] (58.00,-14.75) -- (86.20,-14.75);
\draw[black!45,line width=0.35pt] (58.00,-21.30) -- (86.20,-21.30);
\draw[black!45,line width=0.35pt] (58.00,-21.30) -- ++(0,-0.5);
\draw[black!45,line width=0.35pt] (67.40,-21.30) -- ++(0,-0.5);
\draw[black!45,line width=0.35pt] (76.80,-21.30) -- ++(0,-0.5);
\draw[black!45,line width=0.35pt] (86.20,-21.30) -- ++(0,-0.5);
\node[bn,anchor=north] at (58.00,-21.90) {0};
\node[bn,anchor=north] at (86.20,-21.90) {3};
\node[bn,anchor=north] at (72.10,-21.90) {time (s)};
\draw[pblue,line width=0.75pt,line join=round] plot coordinates {(58.00,-17.63) (58.19,-17.89) (58.38,-18.14) (58.56,-18.38) (58.75,-18.60) (58.94,-18.80) (59.13,-18.97) (59.32,-19.11) (59.50,-19.22) (59.69,-19.29) (59.88,-19.31) (60.07,-19.29) (60.26,-19.22) (60.44,-19.09) (60.63,-18.92) (60.82,-18.69) (61.01,-18.43) (61.20,-18.12) (61.38,-17.78) (61.57,-17.42) (61.76,-17.05) (61.95,-16.68) (62.14,-16.31) (62.32,-15.97) (62.51,-15.65) (62.70,-15.38) (62.89,-15.15) (63.08,-14.97) (63.26,-14.85) (63.45,-14.78) (63.64,-14.78) (63.83,-14.84) (64.02,-14.95) (64.20,-15.11) (64.39,-15.32) (64.58,-15.57) (64.77,-15.85) (64.96,-16.15) (65.14,-16.46) (65.33,-16.79) (65.52,-17.11) (65.71,-17.42) (65.90,-17.71) (66.08,-17.98) (66.27,-18.22) (66.46,-18.44) (66.65,-18.61) (66.84,-18.76) (67.02,-18.87) (67.21,-18.95) (67.40,-18.99) (67.59,-19.00) (67.78,-18.97) (67.96,-18.92) (68.15,-18.84) (68.34,-18.72) (68.53,-18.59) (68.72,-18.42) (68.90,-18.24) (69.09,-18.03) (69.28,-17.80) (69.47,-17.56) (69.66,-17.30) (69.84,-17.03) (70.03,-16.76) (70.22,-16.49) (70.41,-16.23) (70.60,-15.97) (70.78,-15.74) (70.97,-15.54) (71.16,-15.36) (71.35,-15.23) (71.54,-15.13) (71.72,-15.09) (71.91,-15.09) (72.10,-15.14) (72.29,-15.25) (72.48,-15.40) (72.66,-15.60) (72.85,-15.83) (73.04,-16.11) (73.23,-16.41) (73.42,-16.73) (73.60,-17.06) (73.79,-17.39) (73.98,-17.71) (74.17,-18.02) (74.36,-18.30) (74.54,-18.54) (74.73,-18.74) (74.92,-18.89) (75.11,-19.00) (75.30,-19.06) (75.48,-19.07) (75.67,-19.03) (75.86,-18.95) (76.05,-18.83) (76.24,-18.68) (76.42,-18.50) (76.61,-18.30) (76.80,-18.08) (76.99,-17.85) (77.18,-17.62) (77.36,-17.39) (77.55,-17.16) (77.74,-16.94) (77.93,-16.72) (78.12,-16.53) (78.30,-16.34) (78.49,-16.17) (78.68,-16.02) (78.87,-15.88) (79.06,-15.77) (79.24,-15.68) (79.43,-15.61) (79.62,-15.56) (79.81,-15.55) (80.00,-15.57) (80.18,-15.61) (80.37,-15.69) (80.56,-15.80) (80.75,-15.94) (80.94,-16.11) (81.12,-16.31) (81.31,-16.54) (81.50,-16.79) (81.69,-17.06) (81.88,-17.33) (82.06,-17.61) (82.25,-17.88) (82.44,-18.14) (82.63,-18.38) (82.82,-18.59) (83.00,-18.76) (83.19,-18.89) (83.38,-18.97) (83.57,-19.01) (83.76,-18.99) (83.94,-18.93) (84.13,-18.81) (84.32,-18.66) (84.51,-18.47) (84.70,-18.25) (84.88,-18.01) (85.07,-17.75) (85.26,-17.48) (85.45,-17.22) (85.64,-16.96) (85.82,-16.72) (86.01,-16.49) (86.11,-16.29)};
\draw[porange,line width=0.75pt,dash pattern=on 1.4pt off 0.9pt,line join=round] plot coordinates {(58.00,-18.99) (58.19,-19.41) (58.38,-19.78) (58.56,-20.09) (58.75,-20.32) (58.94,-20.48) (59.13,-20.55) (59.32,-20.53) (59.50,-20.44) (59.69,-20.28) (59.88,-20.06) (60.07,-19.80) (60.26,-19.52) (60.44,-19.22) (60.63,-18.94) (60.82,-18.67) (61.01,-18.42) (61.20,-18.21) (61.38,-18.05) (61.57,-17.92) (61.76,-17.82) (61.95,-17.76) (62.14,-17.72) (62.32,-17.70) (62.51,-17.69) (62.70,-17.67) (62.89,-17.65) (63.08,-17.62) (63.26,-17.57) (63.45,-17.50) (63.64,-17.42) (63.83,-17.33) (64.02,-17.25) (64.20,-17.16) (64.39,-17.10) (64.58,-17.05) (64.77,-17.05) (64.96,-17.08) (65.14,-17.17) (65.33,-17.30) (65.52,-17.49) (65.71,-17.72) (65.90,-18.01) (66.08,-18.32) (66.27,-18.66) (66.46,-19.01) (66.65,-19.36) (66.84,-19.68) (67.02,-19.97) (67.21,-20.21) (67.40,-20.38) (67.59,-20.49) (67.78,-20.51) (67.96,-20.46) (68.15,-20.33) (68.34,-20.13) (68.53,-19.87) (68.72,-19.56) (68.90,-19.22) (69.09,-18.86) (69.28,-18.50) (69.47,-18.15) (69.66,-17.82) (69.84,-17.54) (70.03,-17.30) (70.22,-17.12) (70.41,-16.99) (70.60,-16.92) (70.78,-16.90) (70.97,-16.93) (71.16,-17.00) (71.35,-17.10) (71.54,-17.22) (71.72,-17.35) (71.91,-17.49) (72.10,-17.62) (72.29,-17.74) (72.48,-17.84) (72.66,-17.93) (72.85,-18.00) (73.04,-18.06) (73.23,-18.10) (73.42,-18.15) (73.60,-18.19) (73.79,-18.25) (73.98,-18.31) (74.17,-18.40) (74.36,-18.51) (74.54,-18.64) (74.73,-18.79) (74.92,-18.96) (75.11,-19.14) (75.30,-19.33) (75.48,-19.51) (75.67,-19.67) (75.86,-19.81) (76.05,-19.91) (76.24,-19.96) (76.42,-19.96) (76.61,-19.91) (76.80,-19.79) (76.99,-19.61) (77.18,-19.38) (77.36,-19.11) (77.55,-18.80) (77.74,-18.46) (77.93,-18.12) (78.12,-17.78) (78.30,-17.46) (78.49,-17.17) (78.68,-16.94) (78.87,-16.75) (79.06,-16.63) (79.24,-16.57) (79.43,-16.58) (79.62,-16.65) (79.81,-16.78) (80.00,-16.95) (80.18,-17.17) (80.37,-17.41) (80.56,-17.67) (80.75,-17.93) (80.94,-18.18) (81.12,-18.41) (81.31,-18.62) (81.50,-18.80) (81.69,-18.93) (81.88,-19.03) (82.06,-19.09) (82.25,-19.13) (82.44,-19.13) (82.63,-19.12) (82.82,-19.10) (83.00,-19.07) (83.19,-19.05) (83.38,-19.03) (83.57,-19.03) (83.76,-19.04) (83.94,-19.06) (84.13,-19.09) (84.32,-19.13) (84.51,-19.16) (84.70,-19.18) (84.88,-19.18) (85.07,-19.15) (85.26,-19.10) (85.45,-19.01) (85.64,-18.88) (85.82,-18.71) (86.01,-18.51) (86.11,-18.29)};
\end{tikzpicture}
\caption{The problem on the WSCC 9-bus system: from the sparse window of its only PMU (machine~2), the graph neural operator (GNO) estimates the trajectories of all machines.}
\label{fig:sys}
\end{figure}

\IEEEPARstart{S}{tate} estimation turns noisy and incomplete measurements into the estimate of the system state used to monitor, protect, and control power systems~\cite{schweppe1970static,zhao2019dse}. As inverter-based resources replace synchronous machines and their inertia, operating conditions change faster and frequency excursions become larger~\cite{milano2018low}. In this setting, dynamic state estimation (DSE) tracks how rotor angles and speeds evolve in time, which quasi-steady monitoring cannot do~\cite{zhao2019dse}. These trajectories are used in transient stability assessment, out-of-step protection, and wide-area damping control~\cite{zhao2021roles,liu2021control}. However, phasor measurement units (PMUs) are installed at only a few generator buses~\cite{netto2022placement}. A practical estimator must therefore be accurate at the machines without a PMU (Fig.~\ref{fig:sys}) and fast enough to run on every new window.

\begin{figure*}[t]
\centering
\begin{tikzpicture}[
  x=1mm,y=1mm,
  >={Stealth[length=1.7mm,width=1.25mm]},
  font=\footnotesize,
  colttl/.style={font=\footnotesize\bfseries,text=#1,inner sep=0pt,anchor=south},
  ttl/.style  ={font=\footnotesize,text=black!70,inner sep=0.6pt,anchor=north west},
  gcard/.style={rounded corners=1.2mm,line width=0.55pt,draw=black!40,fill=black!3},
  ccard/.style={rounded corners=1.2mm,line width=0.55pt,draw=#1,fill=#1!5},
  wcard/.style={rounded corners=1.2mm,line width=0.55pt,draw=#1,fill=white},
  ln/.style   ={->,line width=0.65pt,draw=black!55},
  lnd/.style  ={->,line width=0.65pt,draw=black!55,densely dashed},
  sml/.style  ={font=\footnotesize,text=black!60,inner sep=0.6pt},
  nd/.style   ={circle,inner sep=0pt,minimum size=#1},
  dot/.style  ={circle,inner sep=0pt,minimum size=0.95mm,fill=pblue}]
\node[colttl=black!70] at (16.00,48.0) {Input}; \draw[black!70,line width=0.4pt] (0.50,47.4) -- (31.50,47.4);
\node[colttl=pblue!85!black] at (63.50,48.0) {Graph encoder}; \draw[pblue!85!black,line width=0.4pt] (38.00,47.4) -- (89.00,47.4);
\node[colttl=porange!90!black] at (123.50,48.0) {Physics decoder}; \draw[porange!90!black,line width=0.4pt] (96.00,47.4) -- (151.00,47.4);
\node[colttl=black!70] at (167.60,48.0) {Output and loss}; \draw[black!70,line width=0.4pt] (155.20,47.4) -- (180.00,47.4);
\draw[gcard] (0.5,32.4) rectangle (31.5,44.6);
\node[ttl] at (1.4,44.2) {\textbf{PMU samples} $y_\ell$ \textbf{of} $\delta_2$};
\draw[black!25,line width=0.35pt] (1.8,33.4) -- (30.2,33.4);
\foreach \p in {(1.80,35.91),(2.37,35.06),(2.94,34.53),(4.07,34.06),(4.64,34.70),(5.21,35.91),(5.78,37.15),(6.34,38.35),(6.91,39.13),(7.48,39.40),(8.05,38.97),(8.62,38.06),(9.18,36.95),(9.75,35.80),(10.89,34.56),(11.46,34.50),(12.02,34.65),(12.59,35.05),(13.16,35.86),(13.73,36.57),(14.86,38.60),(15.43,38.92),(16.00,38.98),(16.57,38.38),(17.14,37.39),(17.70,36.26),(18.84,34.54),(19.41,34.27),(19.98,34.57),(20.54,35.33),(21.68,36.80),(22.25,37.51),(22.82,37.99),(23.38,38.30),(23.95,38.39),(24.52,38.04),(25.09,37.64),(25.66,36.64),(26.22,35.86),(26.79,34.94),(27.36,34.54),(27.93,34.47),(28.50,34.90),(29.63,36.84)} {\node[dot] at \p {};}
\draw[ln] (16,32.0) -- node[sml,right,xshift=0.7mm]{binning} (16,29.4);
\node[sml,anchor=south] at (7.45,26.6) {$b_i$}; \node[sml,anchor=south] at (17.95,26.6) {$\mu_i$}; \node[sml,anchor=south] at (27.45,26.6) {static};
\node[sml,anchor=east,text=black!60] at (3.0,25.20) {1};
\draw[draw=black!35,line width=0.3pt,fill=white] (3.60,24.10) rectangle ++(1.7,2.2); \draw[draw=black!35,line width=0.3pt,fill=white] (14.10,24.10) rectangle ++(1.7,2.2);
\draw[draw=black!35,line width=0.3pt,fill=white] (5.60,24.10) rectangle ++(1.7,2.2); \draw[draw=black!35,line width=0.3pt,fill=white] (16.10,24.10) rectangle ++(1.7,2.2);
\draw[draw=black!35,line width=0.3pt,fill=white] (7.60,24.10) rectangle ++(1.7,2.2); \draw[draw=black!35,line width=0.3pt,fill=white] (18.10,24.10) rectangle ++(1.7,2.2);
\draw[draw=black!35,line width=0.3pt,fill=white] (9.60,24.10) rectangle ++(1.7,2.2); \draw[draw=black!35,line width=0.3pt,fill=white] (20.10,24.10) rectangle ++(1.7,2.2);
\node[sml,text=black!45] at (12.5,25.20) {$\cdots$}; \node[sml,text=black!45] at (23.0,25.20) {$\cdots$};
\draw[draw=pgrey!70,line width=0.3pt,fill=pgrey!75] (24.60,24.10) rectangle ++(1.7,2.2);
\draw[draw=pgrey!70,line width=0.3pt,fill=pgrey!75] (26.60,24.10) rectangle ++(1.7,2.2);
\draw[draw=pgrey!70,line width=0.3pt,fill=pgrey!35] (28.60,24.10) rectangle ++(1.7,2.2);
\node[sml,anchor=east,text=pblue!85!black] at (3.0,22.50) {2};
\draw[draw=pblue!70,line width=0.3pt,fill=pblue!30] (3.60,21.40) rectangle ++(1.7,2.2);
\draw[draw=pblue!70,line width=0.3pt,fill=pblue!75] (14.10,21.40) rectangle ++(1.7,2.2);
\draw[draw=pblue!70,line width=0.3pt,fill=pblue!18] (5.60,21.40) rectangle ++(1.7,2.2);
\draw[draw=pblue!70,line width=0.3pt,fill=pblue!75] (16.10,21.40) rectangle ++(1.7,2.2);
\draw[draw=pblue!70,line width=0.3pt,fill=pblue!15] (7.60,21.40) rectangle ++(1.7,2.2);
\draw[draw=pblue!70,line width=0.3pt,fill=pblue!75] (18.10,21.40) rectangle ++(1.7,2.2);
\draw[draw=pblue!70,line width=0.3pt,fill=pblue!32] (9.60,21.40) rectangle ++(1.7,2.2);
\draw[draw=pblue!70,line width=0.3pt,fill=pblue!75] (20.10,21.40) rectangle ++(1.7,2.2);
\node[sml,text=black!45] at (12.5,22.50) {$\cdots$}; \node[sml,text=black!45] at (23.0,22.50) {$\cdots$};
\draw[draw=pblue!70,line width=0.3pt,fill=white] (24.60,21.40) rectangle ++(1.7,2.2);
\draw[draw=pblue!70,line width=0.3pt,fill=pblue!31] (26.60,21.40) rectangle ++(1.7,2.2);
\draw[draw=pblue!70,line width=0.3pt,fill=pblue!54] (28.60,21.40) rectangle ++(1.7,2.2);
\node[sml,anchor=east,text=porange!90!black] at (3.0,19.80) {3};
\draw[draw=porange!70,densely dashed,line width=0.3pt,fill=white] (3.60,18.70) rectangle ++(1.7,2.2); \draw[draw=porange!70,densely dashed,line width=0.3pt,fill=white] (14.10,18.70) rectangle ++(1.7,2.2);
\draw[draw=porange!70,densely dashed,line width=0.3pt,fill=white] (5.60,18.70) rectangle ++(1.7,2.2); \draw[draw=porange!70,densely dashed,line width=0.3pt,fill=white] (16.10,18.70) rectangle ++(1.7,2.2);
\draw[draw=porange!70,densely dashed,line width=0.3pt,fill=white] (7.60,18.70) rectangle ++(1.7,2.2); \draw[draw=porange!70,densely dashed,line width=0.3pt,fill=white] (18.10,18.70) rectangle ++(1.7,2.2);
\draw[draw=porange!70,densely dashed,line width=0.3pt,fill=white] (9.60,18.70) rectangle ++(1.7,2.2); \draw[draw=porange!70,densely dashed,line width=0.3pt,fill=white] (20.10,18.70) rectangle ++(1.7,2.2);
\node[sml,text=black!45] at (12.5,19.80) {$\cdots$}; \node[sml,text=black!45] at (23.0,19.80) {$\cdots$};
\draw[draw=porange!70,line width=0.3pt,fill=white] (24.60,18.70) rectangle ++(1.7,2.2);
\draw[draw=porange!70,line width=0.3pt,fill=porange!22] (26.60,18.70) rectangle ++(1.7,2.2);
\draw[draw=porange!70,line width=0.3pt,fill=porange!75] (28.60,18.70) rectangle ++(1.7,2.2);
\node[sml,anchor=north] at (16,17.9) {$\mathbf{a}_i\in\mathbb{R}^{2N_b+3}$, $N_b{=}30$};
\draw[ln] (30.8,22.5) -- (33.2,22.5);
\node[draw=pblue!70,fill=pblue!8,rounded corners=0.8pt,line width=0.5pt,minimum width=3.4mm,minimum height=7.4mm,inner sep=0pt] at (35.1,22.5) {\rotatebox{90}{\footnotesize linear}};
\draw[gcard] (0.5,1.2) rectangle (31.5,10.8);
\node[ttl] at (1.4,10.4) {\textbf{Query time}};
\node at (16,4.6) {$\tau=t/T\in[0,1]$};
\foreach \k/\op in {2/0.30,1/0.55,0/1}
  {\begin{scope}[shift={(\k*3.2,\k*2.6)}]
     \fill[pblue!6,opacity=\op] (39.5,17.5) -- (79,17.5) -- (83,35.5) -- (43.5,35.5) -- cycle;
     \draw[pblue!60,line width=0.45pt,opacity=\op] (39.5,17.5) -- (79,17.5) -- (83,35.5) -- (43.5,35.5) -- cycle;
   \end{scope}}
\draw[ln] (37.0,22.5) -- (40.5,22.5);
\node[sml,text=pblue!75!black,anchor=south east] at (89.4,41.3) {$L_g{=}3$ GATv2 layers, each with a skip};
\coordinate (g2) at (49.5,29); \coordinate (g3) at (73,29); \coordinate (g1) at (61.5,21.3);
\draw[black!30,line width=1.7pt] (g1) -- (g2); \draw[black!30,line width=1.37pt] (g1) -- (g3);
\draw[black!30,line width=1.22pt] (g2) -- (g3);
\draw[->,pblue,line width=0.75pt,>={Stealth[length=1.5mm,width=1.15mm]}] (g2) to[bend left=32] (g3);
\draw[->,pblue!70,line width=0.6pt,>={Stealth[length=1.3mm,width=1mm]}] (g2) to[bend right=24] (g1);
\node[nd=3.4mm,draw=pblue,fill=pblue!22,font=\footnotesize] at (g2) {2};
\node[nd=3.4mm,draw=porange,densely dashed,fill=porange!18,font=\footnotesize] at (g3) {3};
\node[nd=3.4mm,draw=black!50,fill=black!12,font=\footnotesize] at (g1) {1};
\node[sml,text=pblue!80!black] at (61.2,34.3) {attention $\alpha_{ij}$, $4$ heads};
\node[sml,anchor=north] at (61.5,16.6) {edges $e_{ij}=\big(\log|Y_{ij}|,G_{ij},B_{ij}\big)$};
\foreach \r/\c in {0/black!45,1/pblue,2/porange}
  {\foreach \j in {0,...,5}
     {\fill[\c,opacity={0.25+0.12*mod(\j+\r,4)}] (90.6,{30.6-5.2*\r+0.62*\j}) rectangle ++(1.5,0.55);}
   \draw[\c,line width=0.35pt] (90.6,{30.6-5.2*\r}) rectangle ++(1.5,3.72);}
\node[sml,anchor=north east] at (95.4,19.6) {$z_i\in\mathbb{R}^{96}$};
\draw[ccard=pviolet] (38.5,1.2) rectangle (88,10.8);
\node[ttl,text=pviolet] at (39.4,10.4) {\textbf{Trunk}};
\draw[pviolet!100,line width=0.5pt,domain=0:1,samples=40,smooth] plot ({40.6+19*\x},{6.60+0.45*sin(1*180*\x)});
\draw[pviolet!85,line width=0.5pt,domain=0:1,samples=70,smooth] plot ({40.6+19*\x},{5.25+0.45*sin(2*180*\x)});
\draw[pviolet!70,line width=0.5pt,domain=0:1,samples=100,smooth] plot ({40.6+19*\x},{3.90+0.45*sin(4*180*\x)});
\draw[pviolet!55,line width=0.5pt,domain=0:1,samples=130,smooth] plot ({40.6+19*\x},{2.55+0.45*sin(8*180*\x)});
\node[anchor=west] at (60.4,5.0) {$\psi(\tau)$, $K{=}8$};
\draw[ln] (76.4,5.0) -- (79.9,5.0);
\node[nd=0.9mm,fill=pviolet!80] (t8100) at (81.00,3.80) {};
\node[nd=0.9mm,fill=pviolet!80] (t8101) at (81.00,5.00) {};
\node[nd=0.9mm,fill=pviolet!80] (t8102) at (81.00,6.20) {};
\node[nd=0.9mm,fill=pviolet!80] (t8340) at (83.40,3.20) {};
\node[nd=0.9mm,fill=pviolet!80] (t8341) at (83.40,4.40) {};
\node[nd=0.9mm,fill=pviolet!80] (t8342) at (83.40,5.60) {};
\node[nd=0.9mm,fill=pviolet!80] (t8343) at (83.40,6.80) {};
\node[nd=0.9mm,fill=pviolet!80] (t8580) at (85.80,3.80) {};
\node[nd=0.9mm,fill=pviolet!80] (t8581) at (85.80,5.00) {};
\node[nd=0.9mm,fill=pviolet!80] (t8582) at (85.80,6.20) {};
\foreach \a in {0,1,2} {\foreach \b in {0,...,3} {\draw[pviolet!35,line width=0.25pt] (t810\a) -- (t834\b);}}
\foreach \a in {0,...,3} {\foreach \b in {0,1,2} {\draw[pviolet!35,line width=0.25pt] (t834\a) -- (t858\b);}}
\draw[ln] (31.6,6) -- (38.2,6);
\begin{scope}[on background layer]
  \draw[rounded corners=2mm,draw=porange!70,fill=porange!5,line width=0.6pt] (96,-0.4) rectangle (151,45.2);
\end{scope}
\draw[wcard=porange] (98.2,23.4) rectangle (117.2,41.6);
\node[ttl,text=porange!90!black] at (99.0,41.2) {\textbf{Initial state}};
\foreach \cx/\cn/\n in {102.2/A/3,106.9/B/4,111.6/C/2}
  {\foreach \r in {1,...,\n} {\node[nd=1.25mm,fill=porange!75!black] (h\cn-\r) at (\cx,{33.8+1.9*(\r-(\n+1)/2)}) {};}}
\foreach \r in {1,2,3} {\foreach \q in {1,...,4} {\draw[porange!45,line width=0.3pt] (hA-\r) -- (hB-\q);}}
\foreach \q in {1,...,4} {\foreach \r in {1,2} {\draw[porange!45,line width=0.3pt] (hB-\q) -- (hC-\r);}}
\node[sml,anchor=west,inner sep=0.2pt] at ([xshift=0.5mm]hC-2.east) {$\hat\delta$};
\node[sml,anchor=west,inner sep=0.2pt] at ([xshift=0.5mm]hC-1.east) {$\hat\omega$};
\node[anchor=north,align=center,inner sep=0pt] at (107.7,29.9) {$\hat x_i(0)=x_i^{\mathrm{eq}}$\\$+\,s\,g(z_i)$};
\draw[wcard=porange] (120.2,23.4) rectangle (148.8,41.6);
\node[ttl,text=porange!90!black] at (121.0,41.2) {\textbf{RK4 unroll}};
\node[sml] at (134.5,36.9) {$\dot{\hat x}=f_{\mathrm{nom}}(\hat x)+(0,c)$};
\foreach \x/\y [count=\i] in {123.2/34.0,127.8/30.4,132.4/28.8,137.0/31.8,141.6/34.6,146.0/33.0}
  {\coordinate (u\i) at (\x,\y);}
\draw[porange!85!black,line width=0.8pt] plot[smooth,tension=0.6] coordinates {(u1) (u2) (u3) (u4) (u5) (u6)};
\foreach \i in {1,...,6} {\node[nd=1.6mm,fill=porange!85!black] at (u\i) {};}
\foreach \i in {2,...,5}
  {\draw[->,porange,densely dashed,line width=0.5pt,>={Stealth[length=1mm,width=0.85mm]}] ([yshift=-2.6mm]u\i) -- ([yshift=-0.9mm]u\i);}
\draw[black!45,line width=0.35pt] (123.2,27.2) -- (123.2,26.4) -- (127.8,26.4) -- (127.8,27.2);
\node[sml,anchor=north west] at (121.0,26.3) {$\Delta t{=}20$\,ms};
\node[sml,text=porange!90!black,anchor=north] at (140.2,26.3) {$+c$};
\draw[ln] (117.2,32.9) -- (120.2,32.9);
\draw[ln] (93.0,28.4) -- node[sml,above]{$z_i$} (98.2,28.4);
\draw[ln] (91.4,16.0) |- (98.0,9.6);
\draw[wcard=porange,densely dashed] (98.2,1.2) rectangle (148.8,12.6);
\node[ttl,text=porange!90!black] at (99.0,12.2) {\textbf{Learned correction} (GNO-soft only)};
\foreach \cx/\cn/\n in {101.6/A/3,105.4/B/4,109.2/C/3}
  {\foreach \r in {1,...,\n} {\node[nd=1mm,fill=porange!75!black] (c\cn-\r) at (\cx,{5.4+1.5*(\r-(\n+1)/2)}) {};}}
\foreach \r in {1,2,3} {\foreach \q in {1,...,4} {\draw[porange!40,line width=0.25pt] (cA-\r) -- (cB-\q);}}
\foreach \q in {1,...,4} {\foreach \r in {1,2,3} {\draw[porange!40,line width=0.25pt] (cB-\q) -- (cC-\r);}}
\node[anchor=west] at (111.0,5.4) {$c_i=s_c\,\mathrm{MLP}\big([z_i,\mathrm{trunk}(\psi)]\big)$};
\draw[lnd] (88.2,6) -- (98,6);
\draw[lnd] (133.2,12.8) -- (133.2,23.2);
\draw[gcard] (155.2,21.6) rectangle (180.0,44.4);
\node[ttl] at (156.0,44.0) {\textbf{All machines}};
\draw[pblue,line width=0.75pt,line join=round] plot coordinates {(157.20,31.45) (157.34,31.00) (157.48,30.57) (157.62,30.16) (157.76,29.78) (157.90,29.43) (158.04,29.13) (158.18,28.88) (158.32,28.70) (158.46,28.58) (158.60,28.54) (158.74,28.58) (158.88,28.70) (159.02,28.92) (159.16,29.22) (159.30,29.61) (159.44,30.08) (159.58,30.61) (159.72,31.19) (159.86,31.82) (160.00,32.46) (160.14,33.11) (160.28,33.74) (160.42,34.34) (160.56,34.89) (160.70,35.37) (160.84,35.77) (160.98,36.07) (161.12,36.29) (161.26,36.39) (161.40,36.40) (161.54,36.30) (161.68,36.11) (161.82,35.83) (161.96,35.47) (162.10,35.04) (162.24,34.55) (162.38,34.03) (162.52,33.48) (162.66,32.92) (162.80,32.37) (162.94,31.83) (163.08,31.32) (163.22,30.85) (163.36,30.43) (163.50,30.06) (163.64,29.75) (163.78,29.50) (163.92,29.31) (164.06,29.17) (164.20,29.10) (164.34,29.09) (164.48,29.13) (164.62,29.22) (164.76,29.37) (164.90,29.56) (165.04,29.80) (165.18,30.08) (165.32,30.40) (165.46,30.76) (165.60,31.16) (165.74,31.58) (165.88,32.03) (166.02,32.49) (166.16,32.96) (166.30,33.43) (166.44,33.89) (166.58,34.33) (166.72,34.73) (166.86,35.09) (167.00,35.39) (167.14,35.63) (167.28,35.79) (167.42,35.87) (167.56,35.86) (167.70,35.77) (167.84,35.59) (167.98,35.33) (168.12,34.98) (168.26,34.57) (168.40,34.10) (168.54,33.58) (168.68,33.02) (168.82,32.45) (168.96,31.87) (169.10,31.31) (169.24,30.78) (169.38,30.30) (169.52,29.88) (169.66,29.53) (169.80,29.26) (169.94,29.08) (170.08,28.98) (170.22,28.96) (170.36,29.03) (170.50,29.17) (170.64,29.37) (170.78,29.64) (170.92,29.95) (171.06,30.30) (171.20,30.68) (171.34,31.08) (171.48,31.48) (171.62,31.88) (171.76,32.28) (171.90,32.66) (172.04,33.03) (172.18,33.37) (172.32,33.69) (172.46,33.99) (172.60,34.25) (172.74,34.49) (172.88,34.69) (173.02,34.85) (173.16,34.97) (173.30,35.04) (173.44,35.06) (173.58,35.04) (173.72,34.96) (173.86,34.82) (174.00,34.63) (174.14,34.39) (174.28,34.09) (174.42,33.74) (174.56,33.34) (174.70,32.91) (174.84,32.45) (174.98,31.98) (175.12,31.49) (175.26,31.02) (175.40,30.57) (175.54,30.16) (175.68,29.80) (175.82,29.50) (175.96,29.28) (176.10,29.13) (176.24,29.07) (176.38,29.10) (176.52,29.21) (176.66,29.40) (176.80,29.67) (176.94,30.00) (177.08,30.38) (177.22,30.81) (177.36,31.25) (177.50,31.71) (177.64,32.17) (177.78,32.62) (177.92,33.04) (178.06,33.43) (178.13,33.78)};
\draw[porange,line width=0.75pt,dash pattern=on 1.4pt off 0.9pt,line join=round] plot coordinates {(157.20,29.10) (157.34,28.37) (157.48,27.73) (157.62,27.20) (157.76,26.79) (157.90,26.52) (158.04,26.40) (158.18,26.43) (158.32,26.59) (158.46,26.87) (158.60,27.24) (158.74,27.69) (158.88,28.18) (159.02,28.69) (159.16,29.19) (159.30,29.66) (159.44,30.08) (159.58,30.44) (159.72,30.74) (159.86,30.96) (160.00,31.12) (160.14,31.23) (160.28,31.29) (160.42,31.33) (160.56,31.35) (160.70,31.38) (160.84,31.42) (160.98,31.48) (161.12,31.57) (161.26,31.68) (161.40,31.82) (161.54,31.97) (161.68,32.12) (161.82,32.27) (161.96,32.38) (162.10,32.46) (162.24,32.47) (162.38,32.41) (162.52,32.26) (162.66,32.03) (162.80,31.70) (162.94,31.29) (163.08,30.81) (163.22,30.26) (163.36,29.67) (163.50,29.06) (163.64,28.46) (163.78,27.90) (163.92,27.40) (164.06,26.98) (164.20,26.68) (164.34,26.50) (164.48,26.46) (164.62,26.55) (164.76,26.77) (164.90,27.12) (165.04,27.57) (165.18,28.11) (165.32,28.70) (165.46,29.32) (165.60,29.95) (165.74,30.56) (165.88,31.12) (166.02,31.61) (166.16,32.02) (166.30,32.34) (166.44,32.56) (166.58,32.69) (166.72,32.72) (166.86,32.67) (167.00,32.55) (167.14,32.38) (167.28,32.17) (167.42,31.94) (167.56,31.70) (167.70,31.48) (167.84,31.27) (167.98,31.09) (168.12,30.94) (168.26,30.81) (168.40,30.71) (168.54,30.63) (168.68,30.56) (168.82,30.48) (168.96,30.39) (169.10,30.27) (169.24,30.12) (169.38,29.93) (169.52,29.71) (169.66,29.44) (169.80,29.15) (169.94,28.84) (170.08,28.52) (170.22,28.20) (170.36,27.92) (170.50,27.68) (170.64,27.51) (170.78,27.42) (170.92,27.41) (171.06,27.51) (171.20,27.71) (171.34,28.02) (171.48,28.41) (171.62,28.89) (171.76,29.43) (171.90,30.02) (172.04,30.61) (172.18,31.20) (172.32,31.76) (172.46,32.25) (172.60,32.66) (172.74,32.98) (172.88,33.19) (173.02,33.29) (173.16,33.28) (173.30,33.16) (173.44,32.94) (173.58,32.63) (173.72,32.26) (173.86,31.84) (174.00,31.39) (174.14,30.94) (174.28,30.50) (174.42,30.10) (174.56,29.74) (174.70,29.44) (174.84,29.20) (174.98,29.03) (175.12,28.92) (175.26,28.86) (175.40,28.85) (175.54,28.87) (175.68,28.91) (175.82,28.96) (175.96,29.00) (176.10,29.02) (176.24,29.03) (176.38,29.01) (176.52,28.97) (176.66,28.92) (176.80,28.86) (176.94,28.81) (177.08,28.77) (177.22,28.77) (177.36,28.81) (177.50,28.91) (177.64,29.07) (177.78,29.29) (177.92,29.58) (178.06,29.92) (178.13,30.32)};
\node[sml,text=pblue,anchor=south west] at (157.0,37.7) {$\hat\delta_2$};
\node[sml,text=porange,anchor=south east] at (178.6,37.7) {$\hat\delta_3$};
\draw[black!40,line width=0.35pt] (157.2,25.6) -- (178.2,25.6);
\draw[black!40,line width=0.35pt] (157.20,25.6) -- ++(0,-0.5);
\draw[black!40,line width=0.35pt] (164.20,25.6) -- ++(0,-0.5);
\draw[black!40,line width=0.35pt] (171.20,25.6) -- ++(0,-0.5);
\draw[black!40,line width=0.35pt] (178.20,25.6) -- ++(0,-0.5);
\node[sml,anchor=north] at (157.2,24.9) {0}; \node[sml,anchor=north] at (178.2,24.9) {3}; \node[sml,anchor=north] at (167.7,24.9) {time (s)};
\draw[ln] (148.8,32.6) -- (155.2,32.6);
\draw[ln] (167.6,21.4) -- node[sml,right,xshift=0.6mm]{$\hat x(t_\ell)$} (167.6,17.6);
\draw[gcard] (155.2,0.0) rectangle (180.0,17.4);
\node[ttl] at (156.0,17.0) {\textbf{Residual loss}};
\draw[pblue,line width=0.65pt,domain=0:20.4,samples=60,smooth] plot ({157.1+\x},{10.2+1.7*sin(\x*24)});
\foreach \x/\o in {2.0/1.4,6.4/-1.3,11.0/1.25,15.6/-1.2,19.6/1.1}
  {\pgfmathsetmacro\yc{10.2+1.7*sin(\x*24)}
   \draw[porange,line width=0.55pt] ({157.1+\x},\yc) -- ({157.1+\x},{\yc+\o});
   \node[dot,fill=pblue!70!black] at ({157.1+\x},{\yc+\o}) {};}
\node[sml,text=black!75] at (167.6,5.1) {$\mathcal{L}\propto\sum_\ell(\hat\delta_2-y_\ell)^2$};
\node[sml,font=\footnotesize\itshape] at (167.6,1.7) {no state labels};
\draw[ln,black!35,rounded corners=1mm] (0.5,37.6) -- (-0.6,37.6) -- (-0.6,-2.6) -- node[sml,above,pos=0.36]{PMU samples $y_\ell$ are the only targets} (162,-2.6) -- (162,-0.2);
\draw[->,line width=0.65pt,densely dotted,draw=pblue,rounded corners=1mm] (155.0,15.2) -- (110.0,15.2) -- (110.0,23.2);
\node[sml,text=pblue,anchor=south] at (152.9,15.5) {$\nabla\mathcal{L}$};
\node[sml,text=pblue,anchor=south] at (121.6,15.5) {through the unroll};
\end{tikzpicture}
\caption{The GNO of Section~\ref{sec:method} (settings of Section~\ref{sec:setup}, window of Fig.~\ref{fig:traj}(a)): a GATv2 branch encodes the PMU window; the physics decoder integrates the swing equations. Dashed: GNO-soft only; dotted: training gradient.}
\label{fig:arch}
\end{figure*}
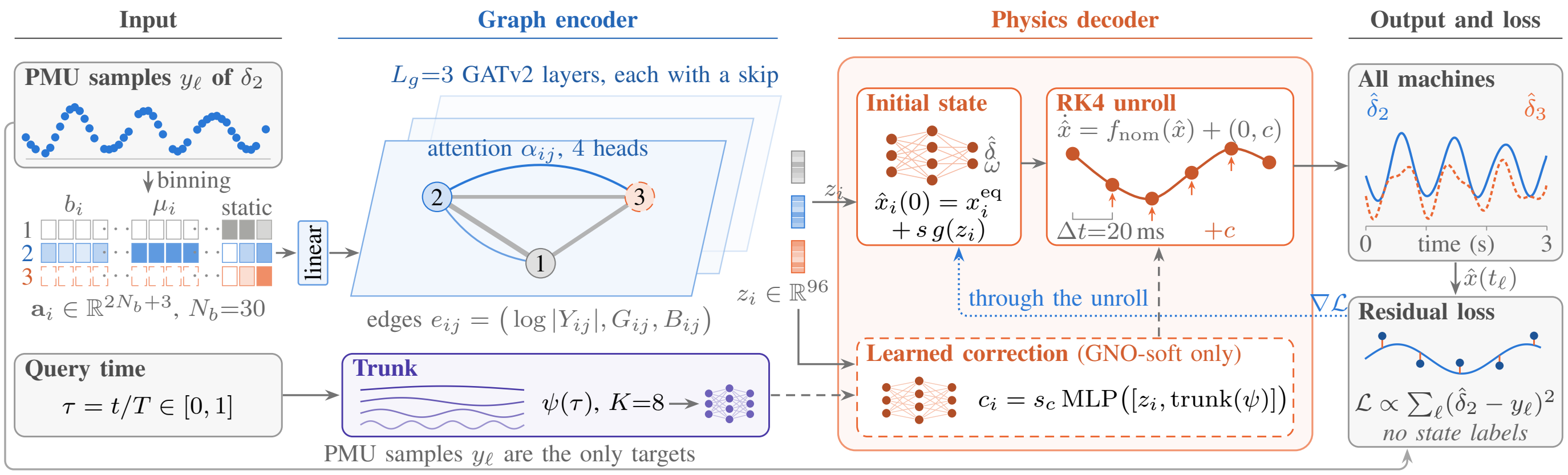

Existing DSE methods fall into two families. First, model-based estimators run a dynamic model forward and correct its state with measurements. The extended and unscented Kalman filters (EKF, UKF) are the most widely used~\cite{zhao2017robust,valverde2011ukf}, also with intermittent PMU data~\cite{paul2018intermittent} and with a Koopman model learned from data~\cite{netto2018koopman}. Moving-horizon estimation (MHE) instead fits a model trajectory to all measurements in a window by nonlinear least squares~\cite{rao2003mhe}. It can recover the internal states of generators that have no PMU~\cite{katanic2022mhe}. However, most of these methods need an accurate model and enough measurements to determine the state. With uncertain parameters or few sensors, their accuracy drops. In addition, MHE must solve a new optimization for every estimate.

Second, data-driven estimators learn a direct map from measurements to states. Since a power network is a graph, graph neural networks (GNNs)~\cite{kipf2017gcn}, which pass information along the topology, are well suited to it. GNNs have been shown to estimate the state even where the PMUs alone cannot determine it~\cite{moshtagh2025gnn}. Similarly, a shallow recurrent decoder (SHRED)~\cite{williams2024shred} reconstructs the full state, unmeasured generators included, from a few PMUs and no system model at run time~\cite{pomarico2026shred}. Beyond synchronous machines, a selective state-space (Mamba) model with expert routing both simulates inverter transients and forecasts them from measurement windows~\cite{wang2026mamba}. However, purely data-driven estimators are trained on state labels, which field data do not provide, and ignore the physical laws that the state must obey.

Physics-informed neural networks (PINNs) combine the two families by adding the governing equations to the learning objective~\cite{raissi2019pinn,huang2023review}. In power systems, PINNs determine rotor angle and frequency, and identify inertia and damping, from few samples of a single-machine infinite-bus system~\cite{misyris2020pinn}, and simulate the differential-algebraic equations of power networks~\cite{moya2023daepinn}. Power-flow equations used as a regularizer help distribution state estimation when few quantities are measured~\cite{ostrometzky2019pinn}. Physics-informed graph learning mainly estimates the bus voltages of distribution feeders~\cite{ngo2023pignn}. However, these works either simulate small systems or estimate static states. Transient stability, in contrast, requires the rotor-angle and speed trajectories of many machines through a disturbance, estimated from a sparse and intermittent measurement window.

This paper addresses this problem. We cast DSE as an operator-learning problem and solve it with a physics-informed graph neural operator (GNO, Fig.~\ref{fig:arch}). A neural operator learns a map between functions, here from a measurement window to a trajectory; examples are deep operator networks (DeepONet)~\cite{lu2021deeponet}, including Bayesian ones for noisy data~\cite{lin2023bdeeponet}, graph kernel operators~\cite{li2020gno}, and physics-informed DeepONets trained from the governing equations~\cite{wang2021pideeponet}. Once trained, it needs only one forward pass per estimate, because the expensive computation is done once, offline (amortized inference~\cite{cranmer2020frontier}). Outside power systems, neural inverse operators solve inverse problems of partial differential equations (PDEs) orders of magnitude faster than optimization~\cite{molinaro2023nio}, and physics-informed variants learn them without labeled solutions~\cite{cho2025pidion}.

Two lines of work are closely related. First, neural operators have served as fast simulators of power-system dynamics in place of an ordinary differential equation (ODE) solver, for post-fault trajectories~\cite{moya2023deeponet} and for machine models~\cite{moya2023generator,karampinis2025no}; however, they map known inputs to trajectories, whereas DSE maps measurements to states. With an implicit Newton layer, a physics-informed operator learns a stiff inverter model without any simulated trajectories~\cite{le2026dae}.  Second, without state labels, per-window MHE~\cite{katanic2022mhe} and Gaussian-process inference with a linearized swing-model prior and most machines metered~\cite{jalali2022gp} recover generators that have no PMU. Both tolerate model-structure mismatch but take the parameters as known, and the latter refits its covariance for each data record. To our knowledge, no learned estimator has recovered such generators without state labels in one forward pass, for nonlinear post-fault trajectories, with a minority of machines instrumented.

We couple the operator to the swing dynamics in two ways. The hard variant, GNO-hard, integrates the nominal model from a learned initial condition, so its output always obeys the model. By contrast, the soft variant, GNO-soft, adds a small, learned forcing term that can absorb model errors. With several PMUs, masked training hides one from the input but keeps its samples in the loss, so that the operator learns machines it does not see. Fig.~\ref{fig:mech} shows how the operator reaches a machine without a sensor, and Fig.~\ref{fig:design} places both variants and the baselines on two design axes. A rotor-angle predictor for transient stability analysis also embeds the rotor equation, corrected by a Runge--Kutta-based error network~\cite{zhang2025hybrid}. In contrast, our operator estimates machines without a sensor and learns from measurements alone. Table~\ref{tab:compare} compares the operator with representative prior work. This paper examines when such an operator, trained without labels, matches per-window fitting at a fraction of its cost, and when its correction or its learned prior allows it to do better.

\begin{table}[t]
\centering
\caption{Positioning against representative prior work (\cmk: yes; $\circ$: partly, see text; --: no).}
\label{tab:compare}
\renewcommand{\arraystretch}{1.35}
\setlength{\tabcolsep}{2.6pt}
\footnotesize

\caption{How machine~3 is reached on the window of Fig.~\ref{fig:traj}(a): (a)~Kron admittances $|Y_{ij}|$ and GNO-hard's attention into machine~3; (b)~machine~2's samples and bins \eqref{eq:bin}, and both machines' trajectories; (c)~$\partial\hat\delta_2(t)/\partial\hat x_3(0)$ through the hard unroll.}
\label{fig:mech}
\end{figure*}

\begin{figure}[t]
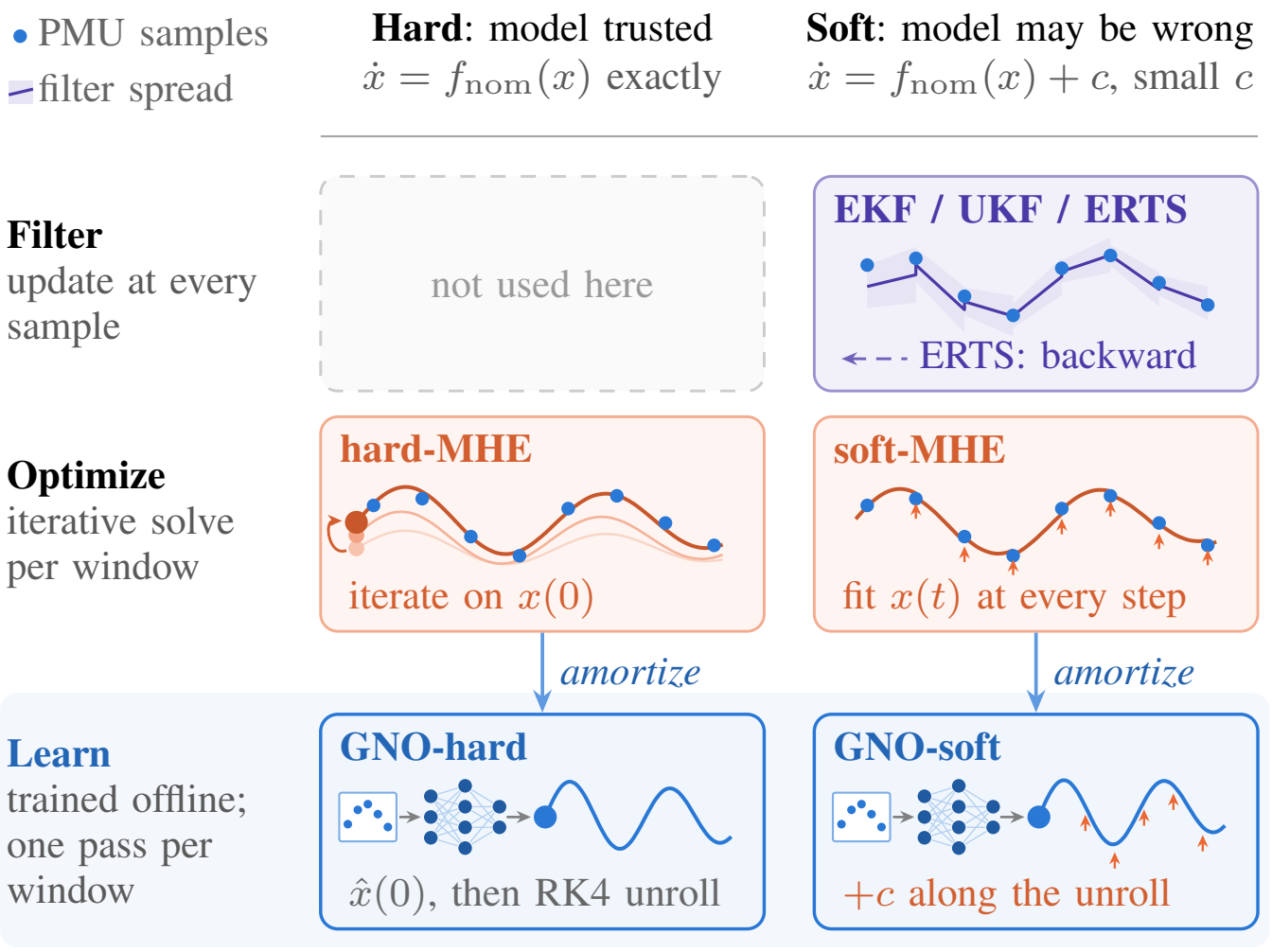

\centering
%
\caption{Estimators by trust in the nominal model (columns) and computation per window (rows). SHRED, trained on state labels, lies outside the grid.}
\label{fig:design}
\end{figure}

To answer this question, we compare against a static baseline, the EKF, the UKF, the EKF's Rauch--Tung--Striebel smoother (ERTS), two MHE variants and the supervised SHRED~\cite{pomarico2026shred}, on the Western Systems Coordinating Council (WSCC) 9-bus, Kundur two-area, IEEE 39-bus and IEEE 118-bus systems. To remove the assumption that rotor angles are measured, we also consider a flux-decay machine with an exciter and an automatic voltage regulator (AVR), observed only through its terminal voltage phasor.

The remainder of this paper is organized as follows. Section~\ref{sec:problem} formulates the problem, and Section~\ref{sec:method} presents the proposed operator. Sections~\ref{sec:baselines} and~\ref{sec:setup} describe the baselines and the experimental setup. Sections~\ref{sec:results}--\ref{sec:realism} report the results, Section~\ref{sec:discuss} discusses them, and Section~\ref{sec:conclusion} concludes. 

\section{Problem Formulation}\label{sec:problem}

\subsection{Multi-Machine Dynamics}
We consider a power network with $N_g$ synchronous machines described by the classical swing model. Let $\delta_i$ be the rotor angle of machine $i$ and $\omega_i$ its per-unit (p.u.) speed deviation, both relative to a reference machine (index $\text{ref}$). Machine $i$ has inertia constant $H_i$ (with $M_i=2H_i$), damping $D_i$, mechanical power $P_{m,i}$, and internal electromotive force (EMF) $E_i$. With synchronous speed $\omega_s$, the dynamics are
\begin{equation}\label{eq:swing}
\dot{\delta}_i = \omega_s\,\omega_i, \qquad
M_i\,\dot{\omega}_i = P_{m,i} - P_{e,i}(\boldsymbol{\delta}) - D_i\,\omega_i,
\end{equation}
where the electrical power output, computed through the reduced admittance matrix $Y=G+jB$, is
\begin{equation}\label{eq:pe}
P_{e,i}(\boldsymbol{\delta}) = E_i\!\sum_{j=1}^{N_g}\! E_j\big(G_{ij}\cos\delta_{ij} + B_{ij}\sin\delta_{ij}\big),
\end{equation}
with $\delta_{ij}=\delta_i-\delta_j$, in a synchronously rotating frame. Throughout, the reference machine is an infinite bus, $\dot{\delta}_{\text{ref}}\equiv\dot{\omega}_{\text{ref}}\equiv0$, so \eqref{eq:swing} is integrated for the other $N_g-1$ machines, a modeling simplification that every estimator here shares. Kron reduction eliminates the network buses (block $\mathrm{n}$) from the augmented admittance matrix. Only the machine nodes (block $\mathrm{g}$) remain, coupled through the topology:
\begin{equation}\label{eq:kron}
Y = Y_{\mathrm{gg}} - Y_{\mathrm{gn}}\,Y_{\mathrm{nn}}^{-1}\,Y_{\mathrm{ng}}.
\end{equation}
A three-phase ($3\phi$) fault changes $Y_{\mathrm{nn}}$ while it lasts and excites transients after clearing (fault buses of the 9-bus system: Fig.~\ref{fig:sys}; its reduced machine graph: Fig.~\ref{fig:mech}(a)).

Stacking $x=[\boldsymbol{\delta}^\top,\boldsymbol{\omega}^\top]^\top\in\R^{2N_g}$ gives a nonlinear ODE $\dot{x}=f(x)$. In practice the true parameters, and hence $f$, differ from the nominal model $f_{\text{nom}}$ that an estimator uses. Besides $H_i$, $D_i$, and $E_i$, the direct-axis transient reactance $X'_{d,i}$ acts through the network: it sets the Kron-reduced admittance, and $E_i$ is defined behind it. We collect these parameters of all machines in $\theta$ and set the mismatch by a relative model error $\varepsilon$,
\begin{equation}\label{eq:eps}
\theta = \theta_{\text{nom}}\odot(1+\varepsilon u),\qquad
u\sim\mathrm{Unif}[-1,1]^{\dim\theta},
\end{equation}
where $\odot$ is the elementwise product and $\mathrm{Unif}$ the uniform distribution. Each perturbed model starts in equilibrium at the pre-fault internal angles $\boldsymbol{\delta}^{\text{eq}}$: $P_{m}=P_{e}(\boldsymbol{\delta}^{\text{eq}})$ is solved again under $\theta$, with \eqref{eq:kron} recomputed if $X'_{d}$ changes, so the mismatch also reaches $P_m$, by an amount comparable to $\varepsilon$ on average on the 9-bus, Kundur, and 39-bus systems.

\subsection{Sparse Intermittent Measurements}\label{sec:meas}
Only a subset $\mathcal{M}\subseteq\{1,\dots,N_g\}$ of machines carries a PMU, and each reports intermittently. Over an estimation window of length $T$, the observations form an irregular set
\begin{equation}\label{eq:obs}
\mathcal{Z} = \{(t_\ell, m_\ell, y_\ell)\}_{\ell=1}^{L}, \quad y_\ell = h_{m_\ell}\big(x(t_\ell)\big) + \nu_\ell,
\end{equation}
where sample $\ell$ of the $L$ retained ones is taken at time $t_\ell$ from machine $m_\ell\in\mathcal{M}$, and $\nu_\ell\sim\mathcal{N}(0,\sigma^2 I)$ is Gaussian measurement noise. Samples are dropped independently for each machine, so the retained times differ between machines. We do not model timestamp jitter, communication latency, or out-of-order arrivals. The output map $h_i$ is what the PMU at machine $i$ reports. In all studies except the detailed-model one, it is the rotor angle, $h_i(x)=\delta_i$, which is an idealization; the detailed-model study (Section~\ref{sec:detailed}) uses the complex terminal voltage phasor that a substation PMU actually observes. 

\subsection{Estimation as Operator Learning}
Classical DSE searches, for each scenario, for the trajectory $x(\cdot)$ that agrees with $\mathcal{Z}$ and $f_{\text{nom}}$. We instead learn, once and for all, an operator
\begin{equation}\label{eq:operator}
\mathcal{F}:\ \mathcal{Z}\ \longmapsto\ x(\cdot)\in C\big([0,T];\R^{n_x}\big),
\end{equation}
that maps any admissible observation window to its state trajectory, with $C$ the continuous functions and $n_x=2N_g$ for \eqref{eq:swing} (more for a detailed machine model, Section~\ref{sec:realism}).

The map is not an exact inverse: with noise and partial observation, many trajectories fit a given $\mathcal{Z}$. Trained to minimize the expected measurement residual over the training scenarios (Section~\ref{sec:method}), the operator favors, among them, those typical of the training scenarios; in effect, it learns a prior (Section~\ref{sec:discuss}). It works on the machine graph $\Gcal=(V,\mathcal{E})$, $|V|=N_g$, whose edges are the nonzero off-diagonal entries of $Y$.

\section{Physics-Informed Graph Neural Operator}\label{sec:method}

\subsection{Operator Architecture}
The GNO (Fig.~\ref{fig:arch}) adapts the branch--trunk design of operator learning~\cite{lu2021deeponet}. A branch encodes the sparse, noisy window into one code per machine, $z\in\R^{N_g\times d}$, of latent width $d$. A physics decoder then maps a query time to the state. It integrates the dynamics from an initial condition set by $z$. Finally, a trunk encodes the query time; only the correction of the soft variant uses it.

\subsubsection{Graph-attention branch}
The window \eqref{eq:obs} is split into $N_b$ uniform time bins per node (strip in Fig.~\ref{fig:mech}(b)). Let $\mathcal{Z}_{ik}$ be the samples of node $i$ that fall in bin $k=\lfloor N_b t_\ell/T\rfloor$, and let $n_{ik}=|\mathcal{Z}_{ik}|$. Then
\begin{equation}\label{eq:bin}
b_{ik}=\frac{1}{\max(n_{ik},1)}\sum_{\ell\in\mathcal{Z}_{ik}} y_\ell,\qquad \mu_{ik}=\mathbf{1}\{n_{ik}>0\},
\end{equation}
with $b_i,\mu_i\in\R^{N_b}$ collecting them over the bins. This gives $2N_b$ channels per node, in radians for angles and in p.u. for phasor components, centered on the pre-fault equilibrium. Three static features are appended, giving
$\mathbf{a}_i=\big[\,b_i,\ \mu_i,\ \mathbf{1}\{i=\text{ref}\},\ H_i/\max_j H_j,\ X'_{d,i}/\max_j X'_{d,j}\,\big]\in\R^{2N_b+3}$. For a voltage phasor, the real and imaginary bin means take the place of $b_i$ and $\mu_i$. A linear layer maps $\mathbf{a}_i$ to $\mathbf{h}^{(0)}_i\in\R^{d}$. It is followed by $L_g$ layers of the graph attention network v2 (GATv2)~\cite{brody2022gatv2} over $\Gcal$, each with a skip connection,
\begin{equation}\label{eq:gat}
\mathbf{h}^{(l+1)} = \mathbf{h}^{(l)} + \phi\!\Big(\mathrm{GATv2}\big(\mathbf{h}^{(l)}, \mathcal{E}; e\big)\Big),
\end{equation}
where $\mathbf{h}^{(l)}$ are the node features after layer $l$ and $\phi$ is the sigmoid linear unit (SiLU) activation. The edge features are $e_{ij}=(\log|Y_{ij}|,G_{ij},B_{ij})$. The first is standardized, and the others are scaled by their largest magnitude. The stack outputs the code $z$, and the attention allows each machine to weight its neighbors differently.

\subsubsection{Initial-condition head}
A multilayer perceptron (MLP) head $g=(g_\delta,g_\omega)$ maps $z$ to a per-machine initial state,
\begin{align}\label{eq:ic}
\hat\delta_i(0) &= \delta_i^{\text{eq}} + s_\delta\,g_{\delta,i}(z), &
\hat\omega_i(0) &= s_\omega\,g_{\omega,i}(z),
\end{align}
where $\boldsymbol{\delta}^{\text{eq}}$ is the pre-fault equilibrium and $s_\delta,s_\omega$ are fixed scales. The reference machine is fixed at its known state. This bounded form keeps training stable.

\subsection{Hard Physics Coupling (GNO-hard)}
GNO-hard produces the trajectory estimate $\xhat$ by integrating the nominal dynamics from the learned initial condition:
\begin{equation}\label{eq:hard}
\xhat(t) = \xhat(0) + \int_0^{t} f_{\text{nom}}\big(\xhat(s)\big)\,ds,
\end{equation}
computed by fixed-step fourth-order Runge--Kutta (RK4) steps inside the network; we call this the unroll. Training gradients flow back through it to $\xhat(0)$ (Fig.~\ref{fig:arch}). The output is thus a trajectory of the nominal model, up to integration error. With the reference machine fixed, the trajectory's only free quantities are the $2(N_g-1)$ initial conditions, or $6(N_g-1)$ for the detailed model of Section~\ref{sec:realism}. GNO-hard thus learns, as a function of the window, the fit that hard-MHE solves anew for every scenario (Fig.~\ref{fig:design}).

\subsection{Soft Physics Coupling (GNO-soft)}
When the nominal model is imperfect ($f_{\text{nom}}\neq f$), a trajectory forced to satisfy \eqref{eq:hard} cannot match the true dynamics. GNO-soft therefore adds a learned correction $c_i(\tau;z)$ to the speed equation of each machine. The correction is a continuous function of the normalized time $\tau=t/T$:
\begin{equation}\label{eq:soft}
\dot{\hat{\delta}}_i = \omega_s\hat{\omega}_i,\qquad
\dot{\hat{\omega}}_i = f^{\text{nom}}_{\omega,i}(\hat{x}) + \underbrace{c_i(\tau;z)}_{\text{learned correction}},
\end{equation}
integrated on the same fixed-step grid from the same learned $\xhat(0)$. Over a step of length $\Delta t$, the correction is held constant. Here $f^{\text{nom}}_{\omega,i}=\big(P_{m,i}-P_{e,i}(\hat{\boldsymbol{\delta}})-D_i\hat{\omega}_i\big)/M_i$ is the speed component of the nominal model.

The correction combines the branch code $z$ with a trunk. The trunk lifts the normalized query time $\tau$ through multi-frequency Fourier features
\begin{equation}\label{eq:fourier}
\psi(\tau) = \big[\sin(2^k\pi\tau),\,\cos(2^k\pi\tau)\big]_{k=0}^{K-1},
\end{equation}
and then through a further MLP, $\mathrm{trunk}$. The correction is $c_i(\tau;z)=s_c\,\mathrm{MLP}\big([\,z_i,\ \mathrm{trunk}(\psi(\tau))\,]\big)$ with a small scale $s_c$, evaluated for each machine with its own code $z_i$ and fixed to zero at the reference.

\subsection{Training Objective}\label{sec:train}
The operator is trained without true state trajectories. Instead, the loss compares the predicted trajectory only with the sparse measurements. Over the window's samples \eqref{eq:obs}, GNO-hard minimizes
\begin{equation}\label{eq:losshard}
\Loss_{\text{hard}}=\frac{1}{|\mathcal{Z}|}\!\!\sum_{(t_\ell,m_\ell,y_\ell)\in\mathcal{Z}}\!\!\big\lVert h_{m_\ell}\big(\xhat(t_\ell)\big)-y_\ell\big\rVert^2 .
\end{equation}
The residual passes through the output map $h$, so the same loss serves voltage phasors, in rectangular coordinates.

The unroll couples the machines. In the 9-bus example, the predicted angle of measured machine~2 responds to the initial state of unmeasured machine~3 (Fig.~\ref{fig:mech}(c)). Fitting the measured machines therefore also shapes the unmeasured ones, without labels for their states. In addition, GNO-soft adds a penalty on the learned correction. The data term is scaled by the noise level $\sigma$ so that the two terms are comparable:
\begin{equation}\label{eq:losssoft}
\Loss_{\text{soft}}=\frac{1}{\sigma^{2}}\,\Loss_{\text{hard}}+w_c\,\big\langle \lVert c(\tau;z)\rVert^{2}\big\rangle ,
\end{equation}
where $\langle\cdot\rangle$ averages over machines and grid points. Varying $w_c$ moves GNO-soft between the two columns of Fig.~\ref{fig:design}. As $w_c\!\to\!\infty$, GNO-soft becomes GNO-hard.

\subsection{Masked Training}\label{sec:masking}
The residual \eqref{eq:losshard} reaches an unmeasured machine only through the coupling. As a result, some combinations of the unmeasured initial states barely affect it. To constrain them, when windows carry several PMUs, we hide one measured machine from the branch in a random half of the training windows. Its bins and mask are zeroed, as for an unmeasured machine, but its samples stay in \eqref{eq:losshard}. The operator thus learns, from measurements alone, to predict machines it does not see. The same idea trains image reconstruction without clean references~\cite{yaman2020ssdu} and graph networks that fill in unsensed nodes~\cite{wu2021ignnk}; to our knowledge it has not been used for DSE. Because the PMU placement is redrawn for each window (Section~\ref{sec:multisys}), every machine is a target in some windows.  Masking is used in the 39-bus and 118-bus studies. GNO-gated blends both couplings (Section~\ref{sec:multisys}).

\section{Baseline Estimators}\label{sec:baselines}
The physics-based baselines fill the other cells of Fig.~\ref{fig:design}. The supervised SHRED is introduced in Section~\ref{sec:intro}.

The static baseline, $\hat{x}(t)\equiv x^{\text{eq}}$, holds the pre-fault equilibrium; it ignores the measurements and shows how hard each scenario is.

The extended and unscented Kalman filters run on the reduced state. They propagate the nominal dynamics on a coarse grid and update whenever a sample arrives. Robust variants~\cite{zhao2017robust} add protection against outliers and non-Gaussian noise, which our data do not contain. The process and measurement noise covariances $Q$ and $R$ are diagonal, scaled to the per-step dynamics mismatch and to the measurement noise. Through $Q$, the filter state can leave the nominal trajectory at every update, so the filters belong to the soft column of Fig.~\ref{fig:design}. Each filter's $Q$, and the UKF's sigma-point spread $\alpha$ ($\beta{=}2$, $\kappa{=}0$), take the values most favorable to it on the test scenarios, from a grid that brackets them or is flat beyond them, so our margins are conservative. ERTS adds a Rauch--Tung--Striebel backward pass to the EKF, tuned the same way; like the operator, it uses the whole window.

Hard-MHE solves
\begin{equation}\label{eq:mhehard}
\min_{x(0)}\ \Loss_{\text{hard}}
\quad\text{s.t.}\quad \dot{x}=f_{\text{nom}}(x),
\end{equation}
by Levenberg--Marquardt least squares, starting from the pre-fault equilibrium.\footnote{The implementation carries an isotropic Tikhonov term ($80^\circ$ per angle, $0.04$~p.u.\ per speed). It has no numerical effect: removing it leaves the reported 118-bus result unchanged. It is stated here for completeness.} Each window starts at fault clearing, so no earlier estimate gives an arrival cost; no inequality constraints are used. Hard-MHE is thus MHE~\cite{katanic2022mhe} without process noise.

Soft-MHE, a benchmark, fits the state at every grid point, with the dynamics as a penalty:
\begin{equation}\label{eq:mhesoft}
\min_{x_0,\dots,x_{m-1}}\ \frac{1}{\sigma^{2}}\Loss_{\text{hard}}+w_d\sum_{k}\big\lVert x_{k+1}-\Phi(x_k)\big\rVert^{2},
\end{equation}
over the states $x_k$ at the $m$ grid points of the window. Here $\Phi$ is one step of the nominal integrator and $w_d$ a penalty weight. The problem is solved by sparse trust-region least squares. The slack in the dynamics, the process noise of a standard MHE~\cite{katanic2022mhe}, absorbs model errors, but each scenario needs a large solve, so it is run on the 9-bus system only.

\section{Experimental Setup}\label{sec:setup}

\subsection{Test System and Scenarios}
Experiments use the WSCC 9-bus, 3-machine system~\cite{andersonfouad}, reduced to its three internal machine nodes. Each scenario applies a three-phase fault at a load bus drawn uniformly from $\{5,6,8\}$, through a shunt admittance $y_f\sim\mathrm{Unif}[3,6]$~p.u. The fault starts at $0.1$~s and is cleared at $t_{\text{cl}}\sim\mathrm{Unif}[0.13,0.20]$~s. The $3$~s window opens at clearing. Trajectories are integrated with RK4 and sampled every $10$~ms. Machine~2 is the only measured machine: $y_\ell=\delta_2(t_\ell)+\nu_\ell$ at every other step (every $20$~ms). Each sample is kept with probability $0.85$, and $\nu_\ell\sim\mathcal{N}(0,\sigma^{2})$ with $\sigma=0.01$~rad ($0.57^\circ$). Machine~1 is the known reference, so machine~3 must be inferred. Model error is injected by perturbing $(H,D,E,X'_d)$ of every machine within $\pm\varepsilon$ of nominal. All estimators use the nominal model.

The static baseline is weak: the mean trajectory of the training pool, $\hat{x}(t)=\mathbb{E}\,x(t)$, also uses no measurements, yet lowers its error by a quarter to a half on our test sets.

\subsection{Metrics and Implementation}
With $\bar{\mathcal{M}}$ the unmeasured machines and $m$ grid points per window, a reported root-mean-square error (RMSE) is
\begin{equation}\label{eq:rmse}
\mathrm{RMSE}=\frac{1}{|\bar{\mathcal{M}}|}\sum_{i\in\bar{\mathcal{M}}}
\Big(\tfrac{1}{m}\textstyle\sum_{k}\big(\hat{v}_i(t_k)-v_i(t_k)\big)^{2}
\Big)^{1/2},
\end{equation}
where $v$ is the rotor angle or, separately, the speed, and $t_k$ are the grid points. This mean of per-machine RMSEs differs from a pooled one when the error is spread unevenly across machines. Angles are in degrees and speeds in $10^{-3}$~p.u. Angles stay within $\pm82^\circ$, so PMU angles never wrap. All estimators see the same $N$ test scenarios of each study, and results are averaged over them, with the standard error of the mean (SEM) as their uncertainty. Two estimators differ significantly when a paired Wilcoxon signed-rank test over the scenarios gives $p<0.05$; otherwise they are statistically tied. Tables~\ref{tab:main}, \ref{tab:multi} and~\ref{tab:round2} report the RMSE \eqref{eq:rmse} in degrees, whose SEM is $4$ to $26\%$ of each mean. Bold marks the best entry, an underline the best without per-scenario optimization where it differs, and an asterisk an entry significantly better than every other. The $p$-values are uncorrected (Bonferroni over $328$: $1.5\times10^{-4}$). A GNO entry averages five seeds, others one run; their pooled results stay within $3\%$ (9-bus), $5\%$ (Kundur, 39-bus) and $13\%$ (detailed, 118-bus); the seed spread raises the SEM by up to a third. In Table~\ref{tab:round2} the filters' initial covariance is matched to the scenario spread. Inference time is the wall-clock time to estimate one scenario on one central processing unit (CPU). The operator reads the whole window, so $\hat{x}(t)$ for $t<T$ is a smoothed estimate and only $\hat{x}(T)$ a real-time one; the filters use past samples only, which favors the window-based estimators.

The operator uses latent width $d{=}96$, $L_g{=}3$ GATv2 layers with $4$ attention heads, $N_b{=}30$ time bins, and $K{=}8$ Fourier frequencies. The query time is normalized to $[0,1]$ over the window. The fixed scales are $s_\delta{=}0.5$, $s_\omega{=}0.01$, and $s_c{=}0.05$. Training uses Adam with learning rate $2\times10^{-3}$, cosine annealing, and gradient clipping at $5$. It runs for $18{,}000$ (hard) and $21{,}000$ (soft) steps with batch size $16$, a pool of $3{,}000$ scenarios, and $w_c{=}10^{4}$. Training scenarios perturb the parameters within $\pm8\%$ of nominal, so test points above $\varepsilon=8\%$ lie outside the training range.

\section{Results and Analysis}\label{sec:results}

\begin{table}[t]
\centering
\caption{WSCC 9-bus: angle RMSE (deg) at the unmeasured machine versus model error ($N{=}50$; static: $6.3$ to $7.5$), and inference time.}
\label{tab:main}
\renewcommand{\arraystretch}{1.3}
\setlength{\tabcolsep}{3.4pt}
\footnotesize
\begin{tabular}{@{}l cccccc c@{}}
\toprule
\multirow{2}{*}{Method} & \multicolumn{6}{c}{RMSE of $\delta$ (deg) vs.\ model error $\varepsilon$} & Inference \\
\cmidrule(lr){2-7}
 & $0\%$ & $5\%$ & $7\%$ & $10\%$ & $15\%$ & $20\%$ & time (ms) \\
\midrule
EKF              & 2.93 & 3.05 & 3.23 & 3.11 & 3.61 & 3.74 & 99 \\
UKF              & 2.87 & 3.02 & 3.14 & 3.06 & 3.57 & 3.74 & 167 \\
ERTS             & \underline{0.39} & 0.64 & 0.81 & 0.95 & \textbf{1.37}\sigmark & \textbf{1.72}\sigmark & 97 \\
GNO-hard (ours)  & \underline{0.39} & 1.27 & 1.54 & 2.33 & 3.46 & 4.12 & 110 \\
GNO-soft (ours)  & \underline{0.39} & \textbf{0.59} & \textbf{0.68}\sigmark & \textbf{0.94} & 1.64 & 2.23 & 123 \\
\midrule
Hard-MHE         & \textbf{0.20}\sigmark & 1.24 & 1.50 & 2.28 & 3.47 & 4.16 & 1214 \\
Soft-MHE         & 0.33 & 0.90 & 1.12 & 1.31 & 1.88 & 2.12 & 9117 \\
\bottomrule
\end{tabular}
\end{table}

\begin{figure*}[t]
\centering
\includegraphics{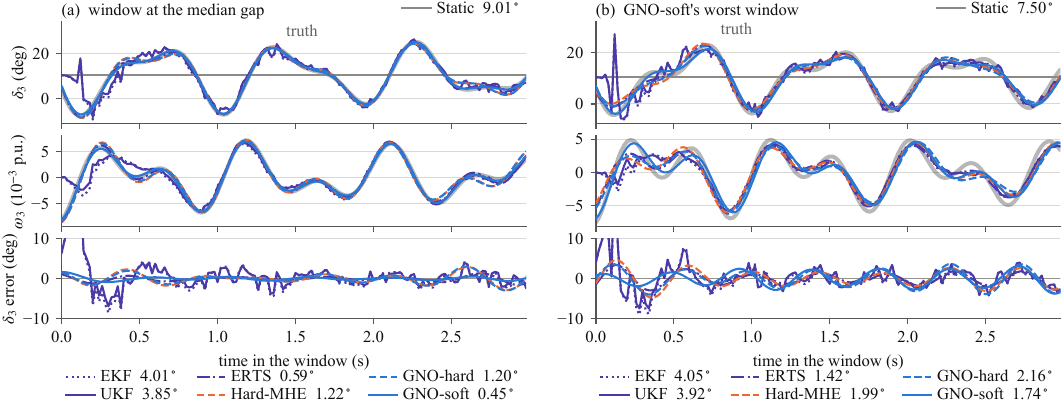}
\caption{Two 9-bus windows at $7\%$: (a)~median gap between hard-MHE and GNO-soft; (b)~GNO-soft's largest error. Grey band: truth; error axis clipped at $\pm10^\circ$; legends: per-window RMSE.}
\label{fig:traj}
\end{figure*}

\subsection{Accuracy Under Model Mismatch}
Fig.~\ref{fig:traj} shows a typical window and the window on which GNO-soft performs worst; even on the latter, it is more accurate than hard-MHE. Table~\ref{tab:main} and Fig.~\ref{fig:ninebus}(a) summarize the results. With a perfect model, hard-constrained fitting is almost exact, and the two operators, which only approximate this fit, are less accurate. Once the model is imperfect, however, the correction of GNO-soft absorbs the error that the hard fit cannot. Up to a model error of $10\%$, GNO-soft and the smoother are the two most accurate estimators and are statistically tied, except at $7\%$, where GNO-soft is better. At larger errors, further beyond the training range, the smoother is better. GNO-hard, by contrast, is on par with hard-MHE at every nonzero model error. On this system, therefore, the gain over hard-MHE comes from the soft coupling rather than from amortization. As Fig.~\ref{fig:mech}(a) suggests, machine~3 attends almost only to the measured machine.

\begin{figure*}[t]
\centering
\includegraphics{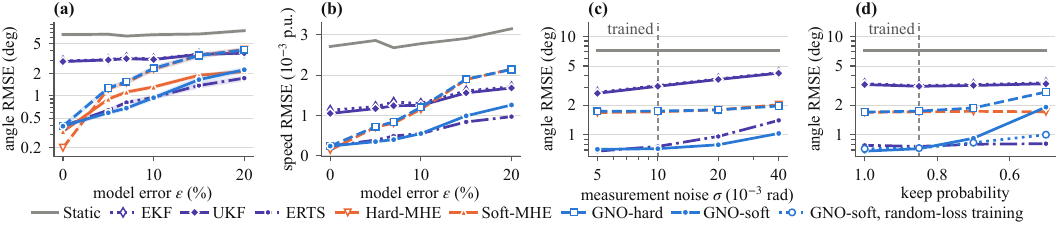}
\caption{WSCC 9-bus. (a)~Angle and (b)~speed versus model error; (c)~noise and (d)~sparsity at $7\%$ (dashed: training value). Bands: $\pm1$ SEM. Dotted: GNO-soft trained with random sample loss (one seed).}
\label{fig:ninebus}
\end{figure*}

\begin{figure*}[t]
\centering
\includegraphics{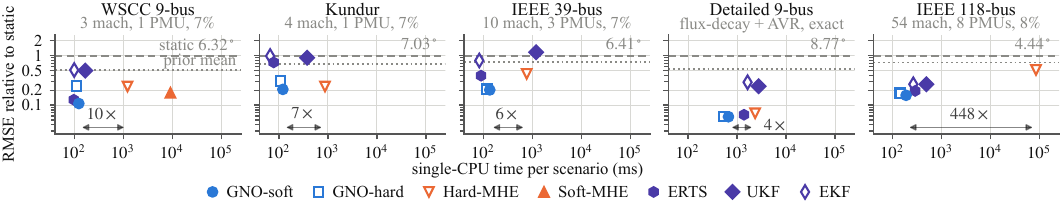}
\caption{RMSE relative to each system's static baseline (value at the dashed line) against single-CPU time per scenario. Dotted: prior mean, no measurements. Spans: hard-MHE to GNO-soft cost ratio.}
\label{fig:cost}
\end{figure*}

\subsection{Soft-MHE and Joint Estimation}
Compared with soft-MHE, GNO-soft is equally accurate with an exact model and at $20\%$ and significantly more accurate from $5$ to $15\%$, at a small fraction of the cost (Fig.~\ref{fig:cost}). By contrast, a joint UKF smoother, whose state is augmented with the errors of the inertias, dampings and EMFs, is significantly more accurate than GNO-soft at every error level, although at a higher cost; knowing which parameters may be wrong thus helps more than a generic correction (Section~\ref{sec:discuss}).

\subsection{Frequency, Noise, and Sparsity}
The rotor speeds rank the estimators as the angles do (Fig.~\ref{fig:ninebus}(b)), except that the hard variants degrade faster and fall behind the UKF at large model errors. Fig.~\ref{fig:ninebus}(c) and (d) vary the noise and the dropout on the same $50$ windows. Because it averages over the whole window, GNO-soft remains the most accurate as the noise grows, whereas the filters degrade the most. Its weakness is a dropout rate not seen in training: when half of the samples are kept, the smoother is significantly more accurate and GNO-soft only ties hard-MHE. Training with random sample loss halves its error there, to below that of hard-MHE.

\section{Architectural Reuse Across Systems}\label{sec:multisys}
Having analyzed the 9-bus system in detail, we retrain the same architecture, with the same hyperparameters, on two more systems. This tests the reuse of the architecture after retraining, rather than the transfer of one trained model to an unseen system. The Kundur system~\cite{kundur1994power}, the standard benchmark for the $0.5$--$0.7$~Hz inter-area mode, has four machines in two areas joined by a weak tie, and one PMU. The IEEE 39-bus system~\cite{athay1979practical} has $10$ machines and three PMUs. On both systems, and on the 118-bus system, the measured machines are redrawn at random among the non-reference machines in every scenario, so the operator must generalize over sensor placements; the 9-bus study keeps one fixed PMU. Each reduced model comes from a solved power flow, and scenarios follow the 9-bus protocol.

\begin{figure}[t]
\centering
\includegraphics{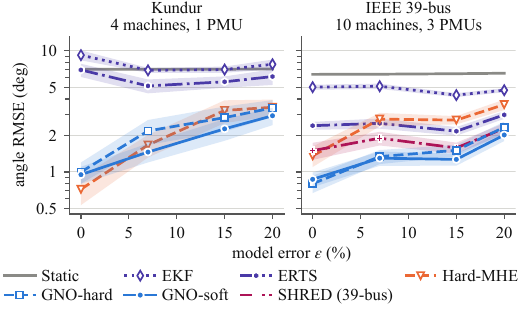}
\caption{Angle RMSE versus model error on Kundur and the 39-bus system (9-bus: Fig.~\ref{fig:ninebus}(a)). UKF omitted (never below $5.8^\circ$). Bands: $\pm1$ SEM.}
\label{fig:multimodel}
\end{figure}

\begin{table}[t]
\centering
\caption{Angle RMSE (deg) at the unmeasured machines versus model error ($N{=}30$; static: $6.4$ to $9.0$ on Kundur, $5.8$ to $7.5$ on the 39-bus), and inference time. SHRED: supervised, untimed.}
\label{tab:multi}
\renewcommand{\arraystretch}{1.1}
\setlength{\tabcolsep}{4pt}
\footnotesize
\begin{tabular}{ll cccc c}
\toprule
System & Method & $0\%$ & $7\%$ & $15\%$ & $20\%$ & ms \\
\midrule
\multirow{6}{*}{\shortstack[l]{Kundur\\{\footnotesize 4-mach}\\{\footnotesize 1 PMU}}}
 & EKF      & 9.21  & 6.90  & 7.00  & 7.74  & 66 \\
 & UKF      & 9.05  & 6.37  & 6.43  & 7.30  & 375 \\
 & ERTS     & 6.95  & 5.13  & 5.55  & 6.13  & 78 \\
 & Hard-MHE & \textbf{0.72}\sigmark & 1.67 & 3.22 & 3.42 & 868 \\
 & GNO-hard & 1.00 & 2.18 & 2.81 & 3.38 & 108 \\
 & GNO-soft & \underline{0.95} & \textbf{1.46} & \textbf{2.27} & \textbf{2.91}\sigmark & 120 \\
\midrule
\multirow{8}{*}{\shortstack[l]{39-bus\\{\footnotesize 10-mach}\\{\footnotesize 3 PMUs}}}
 & EKF      & 5.01  & 5.10  & 4.31  & 4.73  & 82 \\
 & UKF      & 6.79  & 7.52  & 5.88  & 6.24  & 1183 \\
 & ERTS     & 2.41  & 2.52  & 2.17  & 2.96  & 89 \\
 & Hard-MHE & 1.36 & 2.73 & 2.68 & 3.60 & 761 \\
 & SHRED    & 1.50 & 1.90 & 1.58 & 2.29 & -- \\
 & GNO-hard & 0.80 & 1.34 & 1.51 & 2.34 & 117 \\
 & GNO-soft & 0.87 & 1.30 & 1.27 & 2.02 & 134 \\
 & GNO-gated & \textbf{0.75}\sigmark & \textbf{1.28}\sigmark & \textbf{1.27} & \textbf{2.02} & 245 \\
\bottomrule
\end{tabular}
\end{table}

Fig.~\ref{fig:multimodel} and Table~\ref{tab:multi} give the results on the Kundur and 39-bus systems. First, once the model is imperfect, the operator matches or outperforms hard-MHE in a single forward pass. On the 39-bus system, both variants are significantly more accurate than hard-MHE at every nonzero model error, and GNO-hard even with an exact model. On the Kundur system, GNO-soft ties hard-MHE at $7\%$ and is more accurate at larger errors. With an exact model, however, per-window fitting remains the most accurate on the Kundur system, as on the 9-bus system.

\begin{figure}[t]
\centering
\includegraphics{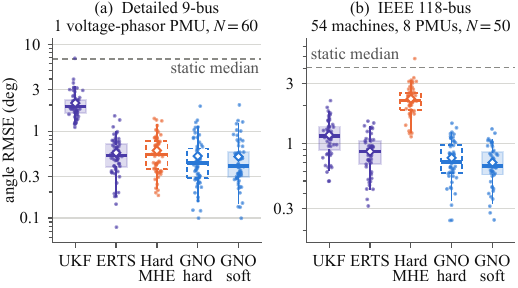}
\caption{Angle RMSE on every scenario of Table~\ref{tab:round2} (log scale). Box: quartiles (dashed: hard coupling); bar: median; diamond: mean; whiskers: $5$th to $95$th percentile.}
\label{fig:round2}
\end{figure}

Second, the recursive filters degrade sharply when few machines are measured. They track the measured machines but leave the unmeasured ones near equilibrium, and even the smoother remains well behind the operator.

Third, the better coupling depends on the model error on every system. With an exact model, a correction can only add variance: the two couplings are tied on the 9-bus system, and the hard coupling is significantly better on the 39-bus system. Under model error, the soft coupling is ahead on all three systems, with a margin that grows with the error on the 9-bus system.\footnote{At $\varepsilon=7\%$, every Kundur GNO-hard run ended well above its best held-out error, which also places it behind GNO-soft with an exact model. We report the final weights of every model.} The learned correction remains small, at most about a tenth of the nominal swing acceleration on the 39-bus system, and increases with the model error. As in multiple-model adaptive estimation~\cite{akhlaghi2019hybrid}, GNO-gated avoids choosing in advance: it blends the soft and hard trajectories with the logistic weight $w=\bigl(1+e^{-(r/\sigma-r_0)/\Delta r}\bigr)^{-1}$ on the soft trajectory and $1-w$ on the hard one, where $r$ is the root-mean-square residual of the hard trajectory at the PMU samples, $\sigma$ the measurement-noise level of \eqref{eq:obs}, and $(r_0,\Delta r)$ are fitted on validation windows, so that $w$ grows with the model error. At the cost of running both operators, GNO-gated is never significantly worse than the better coupling on the 39-bus system, and it is significantly better than both with an exact model and at $7\%$ (Table~\ref{tab:multi}).

Fourth, state labels are beneficial on small systems. A bidirectional long short-term memory (LSTM) network, a Transformer and a plain DeepONet were trained on the simulated states of the pool, which field data do not provide. On the 9-bus and Kundur systems, the best of them is significantly more accurate than GNO-soft at every model error, by up to a factor of four; on the 39-bus system, only at $15$ and $20\%$; and on the 118-bus system, none is. SHRED~\cite{pomarico2026shred}, trained as published on the same data, is significantly less accurate than GNO-soft at every model error.

\section{Detailed Machine Model and Large-System Validation}\label{sec:realism}
This section removes two idealizations made so far, the classical machine model and directly measured rotor angles, and tests a much larger system; the detailed-model and 118-bus operators differ in implementation only: ReLU instead of SiLU, an unnormalized query time with $K{=}6$ (below the Nyquist frequency of the unroll grid), batch size $32$, $w_c{=}10^{3}$, pools of $2000$ and $3000$ scenarios, and $12{,}000$/$18{,}000$ (hard) and $15{,}000$/$21{,}000$ (soft) training steps.

\subsection{Detailed Machine Model with Voltage Phasors}\label{sec:detailed}
We replace the classical model by a one-axis flux-decay machine with an IEEE Type-1 exciter and AVR. Each machine then has the state $x_i=(\delta_i,\omega_i,E'_{q,i},E_{fd,i},V_{R,i},R_{F,i})$, where $E'_{q,i}$ is the $q$-axis transient EMF. The last three states are the field voltage, the regulator output, and its rate feedback. The PMU now reports the terminal voltage phasor (Section~\ref{sec:meas}). Field voltages and regulator setpoints are solved for equilibrium at the power-flow solution, and scenarios follow Section~\ref{sec:setup}. The single measured machine reports its phasor every $20$~ms, with $15\%$ dropout and noise $\sigma=0.01$~p.u.\ per component. This is about $1.4\%$ total vector error, somewhat above the $1\%$ that IEEE Std C37.118.1 allows~\cite{c37118}.

On the detailed model (Table~\ref{tab:round2}, Fig.~\ref{fig:round2}(a)), both operators are more accurate than a UKF driven by the same samples on nearly every scenario. Since the UKF uses past samples only (Section~\ref{sec:setup}), the fair comparison is with the window-based hard-MHE and smoother. GNO-soft is significantly more accurate than hard-MHE and tied with the smoother, and GNO-hard is tied with hard-MHE, both at lower cost. With exact parameters, the two couplings are tied, as on the classical 9-bus model.

\subsection{IEEE 118-Bus System}
To test scale, we move to the IEEE 118-bus system, whose eight PMUs leave about $85\%$ of the $54$ machines unmeasured. Network data and power flow come from MATPOWER~\cite{zimmerman2011matpower}; each of the $54$ machines has $H=4$~s, $X'_d=0.25$~p.u.\ on a base of $1.2$ times its power-flow apparent power (at least $50$~MVA), and $D=2$~p.u.\ on the system base, with constant-admittance loads. Every estimator uses the nominal model, while the scenarios carry $8\%$ parameter error on inertia, damping, and internal EMF.

On this system (Table~\ref{tab:round2}, Fig.~\ref{fig:round2}(b)), GNO-soft is more accurate than the UKF on every scenario and than the smoother on most of them, and its error never exceeds $1.5^\circ$. Masking reduces the error of both operators significantly, by about a tenth, and makes GNO-soft significantly more accurate than GNO-hard. Hard-MHE is less accurate than both operators on every scenario. Its fit has $2(N_g-1)=106$ unknowns, which the eight measured angles reach only through the coupling of an imperfect model; a single window can therefore leave the unmeasured machines poorly determined.

\begin{table}[t]
\centering
\caption{Angle RMSE (deg) at the unmeasured machines and inference time: detailed 9-bus model ($N{=}60$, exact parameters, static $8.77$), IEEE 118-bus system ($N{=}50$, static $4.44$). Asterisks concern the mean.}
\label{tab:round2}
\renewcommand{\arraystretch}{1.0}
\setlength{\tabcolsep}{4pt}
\footnotesize
\begin{tabular}{ll cccc}
\toprule
System & Method & mean & median & 90th pct. & ms \\
\midrule
\multirow{6}{*}{\shortstack[l]{Detailed 9-bus\\{\footnotesize flux-decay\,+\,AVR}\\{\footnotesize 1 voltage PMU}}}
 & EKF      & 2.54 & 2.16 & 3.74  & 1638 \\
 & UKF      & 2.11 & 1.92 & 3.03  & 2752 \\
 & ERTS     & 0.56 & 0.52 & \textbf{0.84} & 1386 \\
 & Hard-MHE & 0.60 & 0.54 & 0.98 & 2339 \\
 & GNO-hard & 0.52 & 0.43 & 0.97 & 540 \\
 & GNO-soft & \textbf{0.51} & \textbf{0.40} & 0.93 & 667 \\
\midrule
\multirow{6}{*}{\shortstack[l]{IEEE 118-bus\\{\footnotesize 54 machines}\\{\footnotesize 8 PMUs}}}
 & EKF      & 1.17 & 1.15 & 1.67  & 264 \\
 & UKF      & 1.17 & 1.15 & 1.67  & 490 \\
 & ERTS     & 0.86 & 0.86 & 1.32  & 288 \\
 & Hard-MHE & 2.28 & 2.20 & 3.13 & 84286 \\
 & GNO-hard & 0.77 & 0.71 & 1.14 & 143 \\
 & GNO-soft & \textbf{0.70}\sigmark & \textbf{0.66} & \textbf{1.03} & 188 \\
\bottomrule
\end{tabular}
\end{table}

\section{Discussion and Limitations}\label{sec:discuss}

\subsection{Why the Operator Helps}
First, graph attention and the coupled unroll (Fig.~\ref{fig:mech}) carry information to an unmeasured machine. Second, the estimate is one (corrected) model trajectory, so samples at all times inform it. Third, training on many scenarios gives the operator a prior over a network's post-fault states. How much each effect matters depends on observability. Where one window determines the state, amortization reduces cost but not error: with two more PMUs in Table~\ref{tab:multi}'s windows and an exact model, hard-MHE ties the operator retrained with them (Fig.~\ref{fig:prior}(a)). More PMUs help hard-MHE far more than the operator, and the prior helps hard-MHE most when they are few. Where it does not, the prior informs the unmeasured machines: given a Gaussian prior fitted to the training states, hard-MHE closes most of the gap on the 118-bus system, less of it on the 39-bus and Kundur systems, and none on the 9-bus one (Fig.~\ref{fig:prior}(b)). To our knowledge, learned DSE methods have so far been compared only with Kalman filters~\cite{tian2020hybrid} or with another learned decoder~\cite{pomarico2026shred}, which cannot separate amortization from the prior.

\begin{figure}[t]
\centering
\includegraphics{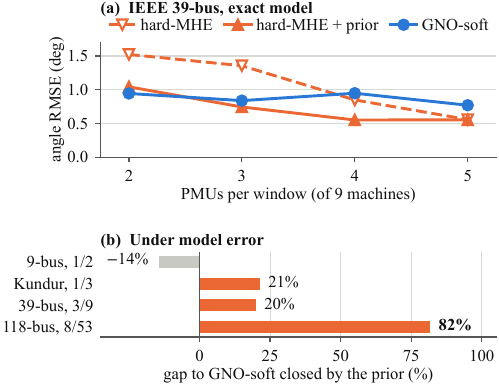}
\caption{Amortization versus the learned prior. (a)~39-bus, exact model; GNO-soft retrained for each $K$. (b)~Nonzero model errors pooled; rows: PMUs/non-reference machines; bars: share of the gap to GNO-soft the prior closes.}
\label{fig:prior}
\end{figure}

\subsection{Ablation and Model Variants}
The ablation study shows that the graph is essential: without message passing the error more than doubles on the 39-bus and 118-bus systems, and a fixed $|Y|$-weighted convolution, a topology-free encoder, or attention pruned to the strongest admittances all cost accuracy, most on the 118-bus system. The physics decoder is equally important: on the 9-bus system a direct trajectory decoder fails, and a swing residual in the loss, at its best weight, stays significantly worse. Removing the edge features costs $11\%$, and $60$ bins cost $5\%$, both significantly; other choices stay within the $3\%$ seed spread.

The operator also works when machine~2 of the 9-bus system reports its terminal-voltage phasor instead of its rotor angle. With exact parameters, GNO-soft ($0.47^\circ$) and GNO-hard ($0.76^\circ$) tie hard-MHE ($0.60^\circ$). Under model error the penalty must be raised; otherwise GNO-soft absorbs the phasor's bias into its correction (with $w_c{=}10^{4}$, $9.4^\circ$ at $7\%$), whereas with $w_c{=}10^{6}$, chosen on validation windows, it is significantly better than hard-MHE from $7$ to $15\%$ ($2.9^\circ$ against $3.6^\circ$ at $7\%$), where the tuned filters do no better than static.

A correction with physical structure outperforms the free one. In this variant, the correction is confined to the swing model: its inertias, dampings and machine couplings are scaled by factors read from $z$, symmetric in each pair, so that it vanishes at the operating point and has no explicit time dependence. On the scenarios of Table~\ref{tab:main}, averaged over three training seeds, it is significantly more accurate than the free correction at every nonzero model error, by $14$ to $24\%$, including the errors beyond the training range. With an exact model, it is slightly less accurate, and it tracks the true mismatch more closely. As with the joint smoother, knowing where the model may be wrong helps more than a generic correction.

\subsection{Complexity and Scaling}
Inference costs $O(L_g N_g^{2} d + N_\tau N_g^{2})$ over $N_\tau$ unroll steps, because Kron reduction makes the machine graph dense. On one CPU thread, the operator costs at most twice as much per window as the EKF and ERTS; for $200$ windows of the 9-bus system processed at once, its forward pass takes $1.6$~ms per window. Training is done once per system: one operator took $42$ to $88$~min on a Tesla T4 GPU for the 9-bus, 39-bus and 118-bus systems, and $3$ to $10$~h on a cluster GPU for the Kundur and detailed models; on the 118-bus system, this equals hard-MHE's CPU time on about $60$ windows. At the end of the window, where every estimate uses past samples only, GNO-soft is never significantly worse than the UKF, and one second of measurements gives it about the accuracy of a three-second window ($0.86^\circ$ over $[0,T]$ against $0.70^\circ$ at $3$~s), whereas half a second does not ($4.03^\circ$).

\subsection{Limitations}
Amortized inference is tuned to the training measurement density (Section~\ref{sec:results}). Masked training needs at least two PMUs per window and varying placements; a machine that is never measured is never a target. Topology errors, false-data injection and model error in the detailed model are not studied. The scale $s_c$ is set by hand and the hard/soft blend fitted offline. The estimates carry no calibrated uncertainty, although the disagreement between three training seeds does rank windows by error (Spearman $\rho{=}0.81$ on the 39-bus system; the tenth of windows with the largest disagreement averages $3.2^\circ$ against $1.2^\circ$ for the rest). Validation uses simulation only. Next steps are inverter-based resources~\cite{le2026dae,wang2026mamba}, hardware-in-the-loop or field PMU data, and governor and stabilizer dynamics.

\section{Conclusion}\label{sec:conclusion}
This paper has presented a physics-informed graph neural operator that maps a sparse, intermittent measurement window to every machine's trajectory and learns from measurements alone. In five settings, with $3$ to $54$ machines, it had the lowest error among estimators that neither optimize per scenario nor use state labels, except on the 9-bus system at large model error, where a tuned Kalman smoother was better. The results suggest that the coupling should follow the model error: hard when the model is exact, and soft otherwise. When the error is unknown, a gate on the hard trajectory's residual can blend the two. Where one window determines the state, amortization saves cost but not error; where it cannot, the operator beat a per-window fit, which we attribute to its learned prior. The code, checkpoints and per-scenario results are available at \url{https://github.com/HuyHoangLe0201/gno-dse}.

\let\oldthebibliography\thebibliography
\renewcommand{\thebibliography}[1]{%
  \oldthebibliography{#1}%
  \setlength{\itemsep}{0pt}%
  \setlength{\parskip}{0pt}}

\end{document}